\documentclass[letterpaper]{article} % DO NOT CHANGE THIS
\usepackage{aaai24}  % DO NOT CHANGE THIS
\usepackage{times}  % DO NOT CHANGE THIS
\usepackage{helvet}  % DO NOT CHANGE THIS
\usepackage{courier}  % DO NOT CHANGE THIS
\usepackage[hyphens]{url}  % DO NOT CHANGE THIS
\usepackage{graphicx} % DO NOT CHANGE THIS
\usepackage{natbib}  % DO NOT CHANGE THIS AND DO NOT ADD ANY OPTIONS TO IT
\usepackage{caption} % DO NOT CHANGE THIS AND DO NOT ADD ANY OPTIONS TO IT
\usepackage{algorithm}
\usepackage{algpseudocode} 
\usepackage{amsmath}
\usepackage[dvipsnames]{xcolor}
\definecolor{pseudogray}{gray}{0.6}

\usepackage{multirow}
\usepackage{booktabs}
\usepackage{arydshln}
    \newcommand{\cdotline}[1]{\noalign{\vskip\abovetopsep}\cdashline{#1}\noalign{\vskip\belowrulesep}}

\usepackage{subcaption}

\usepackage{newfloat}
\usepackage{listings}
\DeclareCaptionStyle{ruled}{labelfont=normalfont,labelsep=colon,strut=off} % DO NOT CHANGE THIS
\floatstyle{ruled}
\newfloat{listing}{tb}{lst}{}
\floatname{listing}{Listing}
\usepackage{etoolbox}
\makeatletter
\pretocmd{\@sect}{\def\@currentlabel{#8}}{}{}% Store title of \section
\pretocmd{\@ssect}{\def\@currentlabel{#5}}{}{}% Store title of \section*
\makeatother

\usepackage[capitalize]{cleveref}
\crefname{section}{Sec.}{Secs.}
\Crefname{section}{Section}{Sections}
\Crefname{table}{Table}{Tables}
\crefname{table}{Tab.}{Tabs.}
\Crefname{equation}{Equation}{Equations}
\crefname{equation}{Eq.}{Eqs.}
\Crefname{algorithm}{Algorithm}{Algorithms}
\crefname{algorithm}{Alg.}{Algs.}

\nocopyright

\usepackage{bibentry}
\begin{document}

\title{Overcoming Overconfidence for Active Learning}
\author{ 
{\normalfont Yujin Hwang\equalcontrib, Won Jo\equalcontrib, Juyoung Hong, and Yukyung Choi}
\\ {\normalfont \normalsize Sejong University}
\\ {\tt\small \{yjhwang, jwon, jyhong, ykchoi \}@rcv.sejong.ac.kr}
}

\maketitle

\begin{abstract}
It is not an exaggeration to say that the recent progress in artificial intelligence technology depends on large-scale and high-quality data. Simultaneously, a prevalent issue exists everywhere: the budget for data labeling is constrained. Active learning is a prominent approach for addressing this issue, where valuable data for labeling is selected through a model and utilized to iteratively adjust the model. However, due to the limited amount of data in each iteration, the model is vulnerable to bias; thus, it is more likely to yield overconfident predictions. In this paper, we present two novel methods to address the problem of overconfidence that arises in the active learning scenario. The first is an augmentation strategy named Cross-Mix-and-Mix~(CMaM), which aims to calibrate the model by expanding the limited training distribution. The second is a selection strategy named Ranked Margin Sampling~(RankedMS), which prevents choosing data that leads to overly confident predictions. Through various experiments and analyses, we are able to demonstrate that our proposals facilitate efficient data selection by alleviating overconfidence, even though they are readily applicable.
\end{abstract}

\section{Introduction} \label{intro}
    The advent of Deep Neural Network (DNN) has facilitated the emergence of an era of large-scale data. Indeed, there is such a growing need for highly informative data that it is regarded with equal significance as the design of improved network architecture~\cite{datadatadata}. The easiest way to envision obtaining this type of data is for domain-specific experts to carefully select instances containing a massive amount of information and then assign appropriate labels to them. However, this strategy is very inefficient when taking into account limited resources such as time and cost.

    \begin{figure}[t!]
        \centering
        \includegraphics[width=0.95\linewidth]{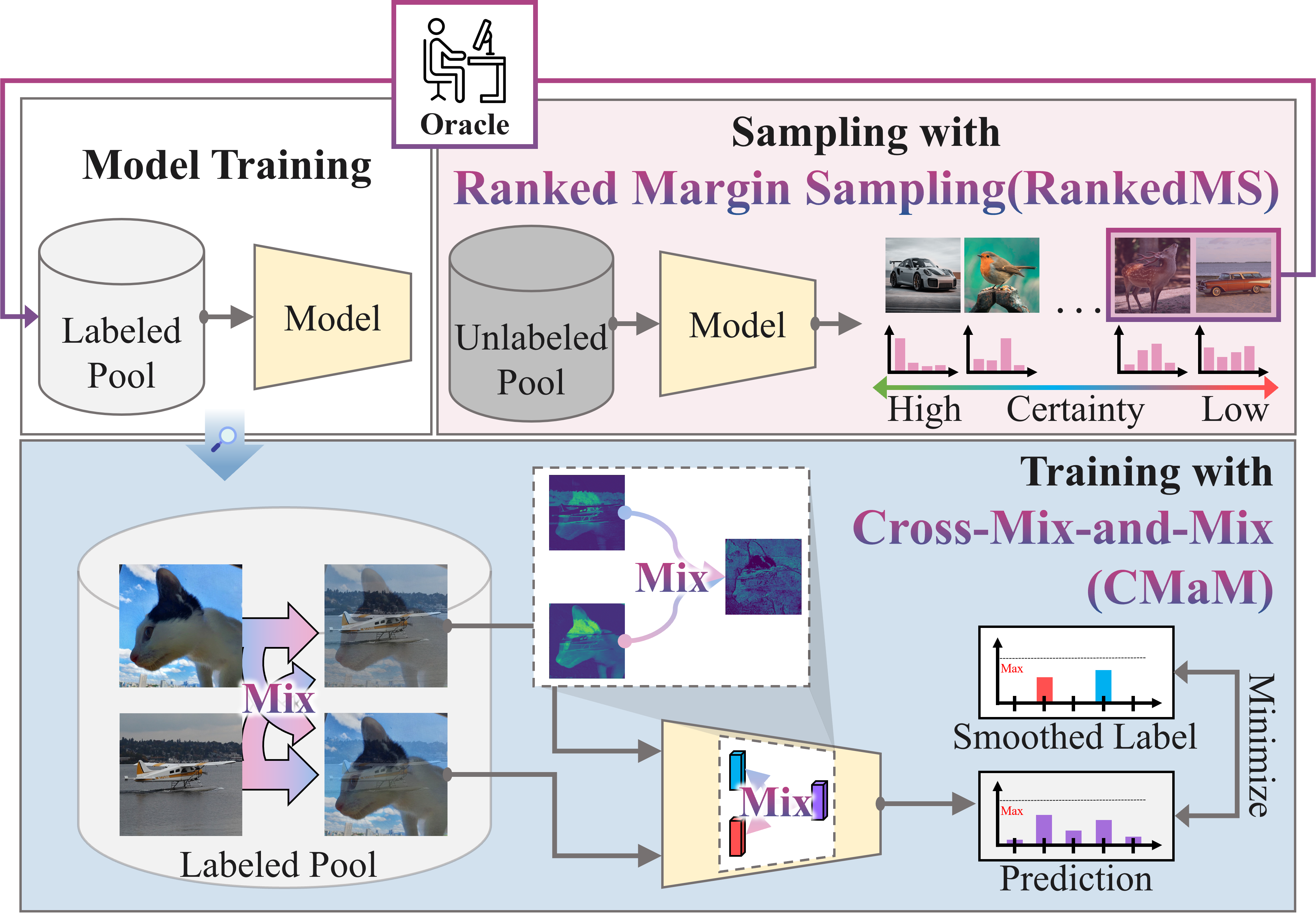} 
        \vspace{-0mm}
        \caption{\textbf{Overview of the Proposed Method.} We propose two novel methods for addressing the problem of bias in a model, which can easily arise in a pool-based active learning scenario. The first is the augmentation technique CMaM for calibrating a model that exhibits overconfidence, and the second is the sampling technique RankedMS for excluding data that causes biased predictions. \vspace{-2mm}} \label{fig:fig1}
    \end{figure}

    Active learning, as a major approach to addressing these issues, has been the subject of ongoing studies. Active learning is a scenario where valuable data is chosen or generated from an unlabeled dataset and a human annotator, referred to as the oracle, is asked to provide manual labeling. In other words, the primary objective of this task is to reduce the cost of labeling while maintaining the quality of learning. Therefore, many studies on active learning have focused on the scenario where small amounts of data are efficiently updated using DNN to reduce cost.
    
    However, it is commonly recognized that when DNN exhibit bias, they have a tendency to show overconfidence in their predictions. Likewise, an active learning model is also easy to overconfident due to bias-inducing factors, including data imbalance, which can occur from the utilization of limited samples for model updates. Furthermore, the iterative process of learning and updating labels can lead to accumulation of bias within the model. As a result, reducing the cost may compromise the quality of the learning process.

    \begin{figure*}[!t]
    \includegraphics[width=\textwidth]{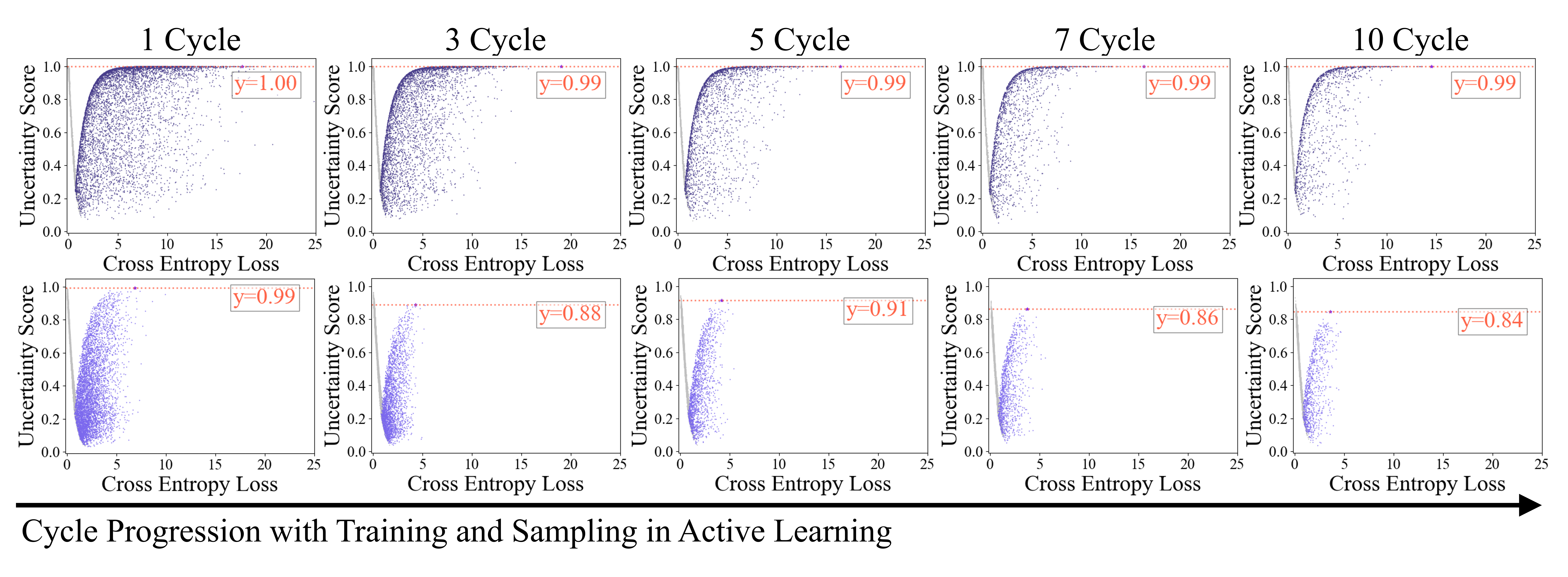} 
        \vspace{-8mm}
        \caption{\textbf{Calibrated Overconfidence Results via the Proposed Method on CIFAR10.} The top row is the result of a random baseline with conventional training and random sampling every cycle. The bottom row is the result of the proposed method. The figures depict predictions for all sample in the test set after training with a labeled dataset in each cycle. The labeled dataset progressively accumulates 1K samples in each cycle (10K samples in the final cycle). In the figures, the x-axis represents the cross entropy loss with GT for the prediction, the y-axis represents the uncertainty score estimated by RankedMS, a gray dot indicates a correct sample, a colored dot indicates an incorrect sample, and the box in the red text indicates the maximum uncertainty score among the wrong samples. If there are many wrong samples with a high margin score, i.e., if there are many samples in the upper right corner of the figures, it indicates that the model is overconfident.
 \vspace{-2mm}} \label{fig:fig2}
    \end{figure*}
    
    In this paper, we present two methods, Cross-Mix-and-Mix (CMaM) and Ranked Margin Sampling (RankedMS), to mitigate overconfidence that arises in the active learning scenario. CMaM is an augmentation applied as a cross for pairs of inputs and intermediate features that calibrates overconfident predictions during the learning process. RankedMS excludes biased data when sampling data by determining how overconfident the prediction for the assessment of value is. \Cref{fig:fig1} depicts an overview of these two methods. Even though both methods are simply applicable, they can help in efficient optimization by preventing the model from becoming overconfident. 
    
    \Cref{fig:fig2} demonstrates the effectiveness of our proposal in reducing overconfidence in an active learning scenario. Due to the fact that only 2\% (i.e., 1K) of all training samples are added per cycle, the random baseline is significantly mispredicted with high probability. In contrast, in the proposed method, a tendency not to overconfidence is observed from the beginning of the cycle, even if samples are incorrectly predicted, suggesting that it prevents the model from being biased. In addition, we provide further experiments and analyses to prove their benefits in later sections.
    
    The following are our main contributions: 
    \begin{enumerate}
        \item We present novel methods, CMaM and RankedMS, to address the challenge of overconfidence in active learning scenarios.
        \item We provide analyses proving that our proposal successfully alleviates the overconfidence issue in the model.
        \item We reveal that the proposed methods demonstrate outstanding effectiveness compared to previous studies conducted on the CIFAR10, CIFAR100, imbalanced CIFAR10, and ImageNet datasets according to previously used experimental settings.
    \end{enumerate}

\section{Related Works}
    \paragraph{Active Learning \\}
        Active learning can be categorized into two streams: query-synthesis-based and pool-based. The former requests labeling after data generation, while the latter requests labeling after the assessment of data usefulness. Given that our work belongs to the pool-based stream, we will briefly address most related works in this stream.
    
        The pool-based stream is commonly separated into two types, the first of which is the diversity-based approach. The diversity-based approach is to select a subset of diverse samples that successfully captures the distribution of the entire unlabeled dataset. Several methods~\cite{cluster, k-means2, cdal, coresets, badge} leverage clustering strategies between features to select diverse samples. In addition, several methods~\cite{distribution2, distribution3, cpal} try to understand the distribution of unlabeled data. Moreover, some methods~\cite{vaal, distribution} employ adversarial network architectures to capture information within the latent space.

        \vspace{0mm}
        The other type of pool-based stream is the uncertainty-based approach. The uncertainty-based approach is to select a subset of uncertain samples depending on the current state of a model. This approach is based on the assumption that the labeling of the data, which the model ambiguously predicts, will contribute to the learning process. The predominant method is to assess the predictions of a model due to its explicit quantification of uncertainty in the model. Entropy-based methods~\cite{ceal, entropy1, entropy2} and margin-based methods~\cite{margin, margin2, margin3} are examples of this strategy. The former utilizes the entropy of predictions for quantification, while the latter utilizes the two most likely probabilities of that. In addition, a variety of strategies have progressively emerged. Some methods~\cite{ical,alfa-mix} attempt to select data by leveraging the consistency of predictions for perturbed versions. Several methods~\cite{sraal, tavaal} employ an adversarial network to figure out the informativeness of unlabeled samples. In addition, several methods~\cite{gcnal, vab-al, mcdal} utilize a auxiliary network to measure uncertainty. Furthermore, a method~\cite{lloss} selects ambiguous data for a model by learning to predict losses itself. However, methods in this category are vulnerable to overconfidence because they do not account for the risk that the model can be biased by a small number of samples. In contrast, our method reduces overconfidence within a limited budget by expanding the training distribution and preventing the selection of data that leads to overconfident predictions. In the experimental section, we demonstrate that our proposal has a benefit for calibrating the model in an active learning scenario.

    \paragraph{Model Calibration \\}
        Most neural network models for the classification task are trained using one-hot encoded labels, where all of the probability distribution is concentrated in a single class. In this scenario, the model has the potential to be trained in a manner that may lead to the induction of overconfidence. Thus, many studies in this area have been conducted with the objective of mitigating overconfidence. One of the studies~\cite{calib1} addressed the calibration of models, finding that the capacity, normalization, and regularization of a model significantly impact the calibration of the network. Mixup~\cite{zhang2017mixup} tried to calibrate the generalization ability of the model by mixing a pair of images and their labels, and subsequent research~\cite{mixupoverconf} has shown that the mixup can help alleviate overconfidence. Several studies addressing regularization~\cite{label_smoothing1, label_smoothing2} sought to solve the problem of overconfidence within the model through label smoothing. Several studies~\cite{calib_imbalanced1, calib_imbalanced2} focused on the issue of imbalanced dataset composition for calibration purposes.

\section{Method}
    \paragraph{Problem Definition \\}
        To begin, we will formalize the scenario of active learning. Let us denote the labeled dataset by~$\mathcal{D}_{L}$ and the unlabeled dataset by~$\mathcal{D}_{U}$. This scenario consists of the iterative process of selecting a subset~$\mathcal{D}_{S}$ with $N_S$~samples and then including~$\mathcal{D}_{S}$ with their labels into~$\mathcal{D}_{L}$. The process continues for $R$~times, where the sampled dataset picked for each cycle~$r \in \left [ \, 1, R \, \right ]$ is denoted by~$\mathcal{D}_{S}^{(r)} $. 

        The careful selection of the most informative~$\mathcal{D}_{S}^{(r)} $ is crucial to this scenario; to this end, the uncertainty-based approach, to which our method belongs, has mainly exploited the ambiguity of the model predictions. Formally, the model~$\mathcal{M}$ can be parameterized by~$\theta$. Furthermore, assuming a classification task, $\mathcal{M}$ can be decomposed into the encoder~$\mathcal{M}_{E}$ and the classifier~$\mathcal{M}_{C}$ with an expression of~$\mathcal{M}$~$=$~$\left \{  \mathcal{M}_{E}, \mathcal{M}_{C} \right \}$, which is also parameterized by~$\theta$~$=$~$\left \{  \theta_{E}, \theta_{C}  \right \}$. Given the input data~$x$, it is projected into a feature space via~$I_E$ layers within~$\mathcal{M}_{E}$, i.e., $z$~$=$~$\mathcal{M}_{E}\left ( x; \theta_{E} \right )$, and then converted to posterior probabilities via~$I_C$ layers within~$\mathcal{M}_{C}$, i.e., $P\left ( y\mid z \right )$~$=$~$\textup{softmax}\left ( \mathcal{M}_{C}\left ( z;\theta_{C} \right ) \right )$. Once the model is trained to minimize the cross entropy loss~$\mathcal{L}$ computed from~$\mathcal{L}(P\left ( y\mid z \right ), y)$, it is common practice to measure the uncertainty using the certainty-aware function~$\phi$. 

        In this process, $\mathcal{D}_{S}^{(r)} $ is likely to be biased because it is generally composed of small samples. This allows $\mathcal{M}$ to easily become overconfident with false results. Therefore, for a successful active learning scenario, $\mathcal{M}$ needs a learning strategy that effectively alleviates overconfident results, and $\phi$ should possess the ability to assess uncertainty while paying attention to overconfidence. In order to address this issue, we will proceed by providing a description of CMaM and RankedMS in the subsequent step.

    \paragraph{Cross-Mix-and-Mix (CMaM) \\} 
        The data augmentation technique called mixup~\cite{zhang2017mixup} was proposed to improve the generalization and robustness of neural network architectures. It linearly interpolates the raw inputs $x_1$, $x_2$, and their one-hot labels $y_1$, $y_2$, by a random variable $\lambda$, as shown in the following equation:
        \begin{align} \label{eq:mixup}
            \dot{x} = \lambda x_1 + (1 - \lambda)x_2  \nonumber
            \\
            \dot{y} = \lambda y_1 + (1 - \lambda)y_2.
        \end{align}
        By expanding the training distribution in this manner, mixup effectively leads to the stabilization of the learning process. Follow-up research~\cite{mixupoverconf} demonstrated that this simple technique is helpful for calibrating neural networks, particularly because interpolated soft labels reduce overconfidence. However, given that they only focus on an abundant setting with a sufficient amount of training data, it is required to further extend the training distribution for calibrating overconfidence in the active learning scenario, where there are much fewer training samples available.

        Here, we present an enhanced training technique, Cross-Mix-and-Mix (CMaM), which can effectively reduce overconfident predictions when given limited training samples. In contrast to mixup, which only uses a linear interpolation of training samples in input space, CMaM can enlarge the training distribution by interpolating in both input space and intermediate feature space. The formulation for our cross-augmentation in two spaces, given randomly sampled inputs $x_1$ and $x_2$, is as follows: 
        \begin{align} \label{eq:cmam}
            \dot{x}_1 &= \lambda_1 x_1 + (1 - \lambda_1)x_2  \nonumber
            \\
            \dot{x}_2 &= (1 - \lambda_1) x_1 + \lambda_1 x_2 \nonumber 
            \\
            \dot{z}^{(I)}_1 &= \mathcal{M}_{E}(\dot{x_1};\left \{ \theta_{E}^{i} \right \}^{I}_{i=1}) \nonumber 
            \\
            \dot{z}^{(I)}_2 &= \mathcal{M}_{E}(\dot{x_2};\left \{ \theta_{E}^{i} \right \}^{I}_{i=1}) \nonumber 
            \\
            \ddot{z}^{(I)} &= \lambda_2 \dot{z}^{(I)}_1 + (1 - \lambda_2) \dot{z}^{(I)}_2,
        \end{align}
                
        \noindent where $\dot{z}^{(I)}_1$ and $\dot{z}^{(I)}_2$ represent intermediate features, outputs of the $I$-th layer of~$\mathcal{M}_{E}$, for interpolated inputs~$\dot{x}_1$ and $\dot{x}_2$, respectively. In addition, $\lambda_1$, $\lambda_2 \in [0,1]$ are independent random variables; thus, one-hot encoded labels $y_1$ and $y_2$ for each input are converted according to the two independent interpolations as follows: 
        \begin{align} \label{eq:y_label}
        \dot{y}_1 &= \lambda_1 y_1 + (1 - \lambda_1)y_2  \nonumber
        \\
        \dot{y}_2 &= (1 - \lambda_1) y_1 + \lambda_1 y_2 \nonumber 
        \\
        \ddot{y} &= \lambda_2 \dot{y}_1 + (1 - \lambda_2) \dot{y}_2 \nonumber
        \\ 
         &= (1 - \lambda_1-\lambda_2 + 2\lambda_1 \lambda_2) y_1 \nonumber
         \\
         & \, \, \, \, \, \, \, \, \, \, \,  + (\lambda_1 + \lambda_2 - 2\lambda_1 \lambda_2) y_2.
        \end{align}
        As a result, the virtual feature~$\ddot{z}^{(I)}$, which is processed by two interpolations, is subsequently embedded across the remaining layers of~$\mathcal{M}_{E}$ and the layers of~$\mathcal{M}_{C}$. Although CMaM can be easily implemented with minimal computational overhead like mixup, it is less probable for the model to yield biased predictions when trained on small samples. This is because twice-augmented data in two spaces is utilized as input in specific locations within~$\mathcal{M}$, thereby offering a more diverse form of input, i.e., an expanded training distribution, to subsequent layers than once-augmented data in a single space. We empirically found that the model exhibits better performance compared to others continuously from the early cycle (i.e., when there are significantly fewer training samples) of the selection process in the active learning scenario. \cref{alg:algorithm} provides further details about the entire training process with CMaM.

        \begin{algorithm}[!t]
        \caption{OO4AL Training Process with CMaM}\label{alg:algorithm}
        \textbf{Inputs}: Labeled dataset~$\mathcal{D}_{L}$, Encoder $\mathcal{M}_{E}(\theta_{E}$), Classifier~$\mathcal{M}_{C}(\theta_{C}$), Cross entropy loss $\mathcal{L}$
        \begin{algorithmic}[1] %[1] enables line numbers

            \For{$e=0,1,2,...$} 
                \Comment{Training up to max epochs}
                \State {$(x_1, x_2, y_1, y_2) \sim \mathcal{D}_{L}$ } \Comment{Sampling w/o replacement}

                \State Compute $\ddot{z}^{(I)}$ by using \cref{eq:cmam}
                \State Compute $\ddot{y}$ by using \cref{eq:y_label}

                % \State {\textcolor{pseudogray}{\# $I_E$: Total number of layers in~$\mathcal{M}_{E}$}}
                \State $\ddot{z} \leftarrow \mathcal{M}_{E}(\ddot{z}^{(I)};\left \{ \theta_{E}^{i} \right \}^{I_E}_{i=I+1})$
                \State $P\left ( \ddot{y}\mid \ddot{z}\right ) \leftarrow \textup{softmax}\left ( \mathcal{M}_{C}\left ( \ddot{z};\theta_{C} \right ) \right )$
                \State Minimize $\mathcal{L}(P\left ( \ddot{y}\mid \ddot{z}\right ), \ddot{y})$
            \EndFor
        \end{algorithmic}
        \end{algorithm}

    \paragraph{Ranked Margin Sampling (RankedMS) \\}
        Observing the prediction of~$\mathcal{M}$ is a typical approach for quantifying its uncertainty. Indeed, it is more explicit to assess the uncertainty in the final outcome rather than attempt to estimate it based on the internal factor of~$\mathcal{M}$. This is because the internal operations of~$\mathcal{M}$ are significantly intertwined with numerous nodes, making it difficult to accurately evaluate the uncertainty depending on a single factor alone. Hence, we introduce a certainty-aware function based on the prediction, Ranked Margin Sampling (RankedMS).

        The objective of RankedMS is to quantify the level of uncertainty in predictions while simultaneously excluding samples that exhibit evidence of overconfidence. The reason for this exclusion is that~$\mathcal{M}$ is already biased for this type of sample, so including them in the next cycle will worsen the generality of~$\mathcal{M}$ even further. To this end, RankedMS is formulated as follows:
        \begin{align} \label{eq:rankedms}
            \phi(P\left ( y\mid x \right )) &= \sum_{{c}'=1}^{\mathcal{C}-1}\frac{P\left ( y^{(c'+1)}\mid x \right ) - P\left ( y^{(c')}\mid x \right )}{c'},
        \end{align}
            where $c'$ represents the sorted index by posterior probabilities, $c'$~$=$~$1$ indicates the highest probability, and $c'$~$=$~$\mathcal{C}$ indicates the lowest probability. For a better comprehension of our key difference from others, suppose three toy predictions: $ P_{a}=\left \{ .1, .1, .7, .1 \right \}$, $P_{b}=\left \{ .3, .23, .23, .24 \right \}$, and $P_{c}=\left \{ .5, .45, .05, .0 \right \}$. Based on intuitive observation, it is apparent that the order of overconfidence is $P_b$, $P_c$, and $P_a$ in ascending order; $P_{b}$ can be the least overconfident prediction because it has similar probabilities over all classes. However, Entropy sampling~\cite{real_entropy}, one of the most typical approaches, selects in the order of $P_b$,~$P_a$,~and~$P_c$ since it does not account for overconfidence. Similarly, Margin sampling~\cite{margin}, another approach, does so in the order of $P_c$, $P_b$, and $P_a$ since it only considers the two highest probabilities. Unlike these, RankedMS yields the same selection order as the intuitive observation above. While the previous example provides one specific case, it is clear that our approach, by virtue of its structure, is basically sensitive to overconfident samples compared to the others. The experimental section demonstrates the effectiveness of considering overconfidence in the quantification of uncertainty, leading to state-of-the-art performance despite its minimal computational cost. \cref{alg:sampling} provides further details about the entire sampling process with RankedMS.

            \vspace{4mm}

        \begin{algorithm}[tb]
        \caption{OO4AL Sampling Process with RankedMS}\label{alg:sampling}
        \textbf{Inputs}: Labeled dataset $\mathcal{D}_{L}$, Unlabeled dataset $\mathcal{D}_{U}$, Selected subset $\mathcal{D}_S$, Budget size $b$
        
        \begin{algorithmic}[1] %[1] enables line numbers
        
        \State {$x_u \sim \mathcal{D}_{U}$ }
        \Comment{$x_u \subset  \mathcal{D}_{U}$}
        \State {$P\left ( y\mid x_u \right ) \gets \mathcal{M}(x_u; \theta)$}    
        \State {Compute $\phi(P\left ( y\mid x_u \right ))$ by using \cref{eq:rankedms}}    
        
        \State{$\mathcal{D}_{S} \gets \underset{(x_u^1, ..., x_u^b)\subset x_u}{\mathrm{argmin}} \phi(P\left ( y\mid x_u \right )) $} 

        \State {$\mathcal{D}_{L} \gets \mathcal{D}_{L} \cup \mathcal{D}_{S}$}
        \State {$\mathcal{D}_{U} \gets \mathcal{D}_{U} \setminus \mathcal{D}_{S}$}
        
        \end{algorithmic}
        \end{algorithm}

\section{Experiments}
    \subsection{Ablation Studies}
        \paragraph{Setup and Implementation Details.}
            Ablation studies were conducted on CIFAR10~\cite{cifar}. The dataset comprises 32x32 images for 10 classes, with a train set of 50K and a test set of 10K. For the active learning scenario, 1K samples are chosen (first cycle: random, other cycles: individual manner) and used for training a model every cycle, for 10 cycles within the train set. ResNet-18~\cite{he2016deep} was employed as the target model, and the hyperparameters for training were consistently set across all experiments. These details can be found in the supplementary material. All experiments were fixed, and the results were reported as the average of five trials with different random seeds on the test set. The evaluation metrics were accuracy and Overconfidence Error~(OE)~\cite{mixupoverconf}. For convenience, The combination of CMaM and RankedMS is referred to as Overcoming Overconfidence for Active Learning (OO4AL).

        \begin{figure}[t!]
            \centering
            \includegraphics[width=0.95\linewidth]{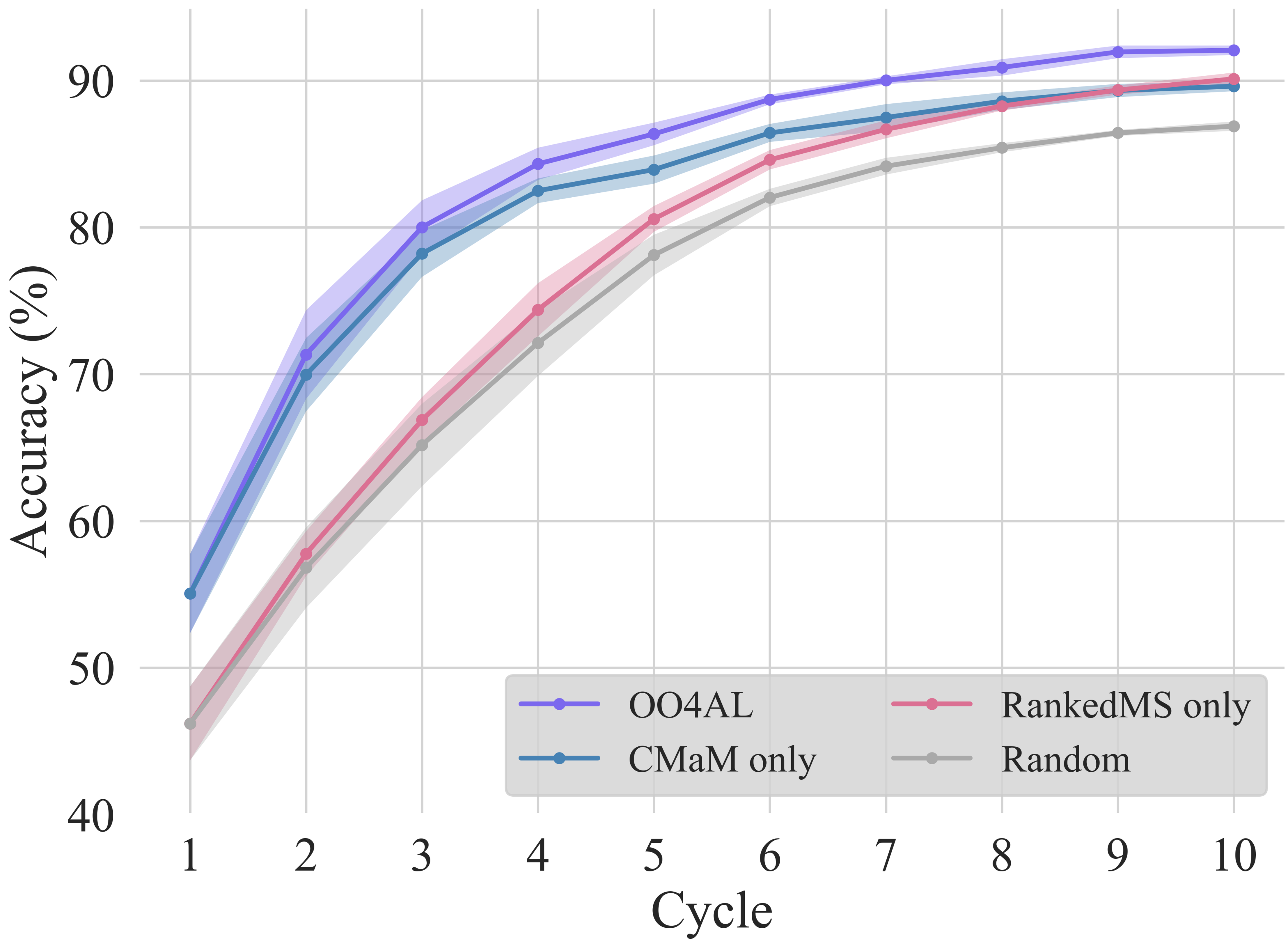} 
            \vspace{-0mm}
            \caption{\textbf{Ablation Study of Proposed Methods.} The number of samples budgeted per cycle is 1K. The accuracy is expressed as the mean and standard deviation of five trials. \vspace{-2mm}} \label{fig:fig_module}
        \end{figure}
        
        \paragraph{Efficiency of the Proposed Methods.}
            This section covers ablation studies conducted for each of the two proposed methods. As shown in~\Cref{fig:fig_module}, both methods exhibit improved results when compared to the random baseline. Specifically, when using only CMaM, the generalization capability of the model can be enhanced by expanding the training distribution. This is demonstrated by the significant boost in performance observed in the early cycles, which have a limited number of samples. When using only RankedMS, it helps the model be less biased than random sampling by selecting samples that do not induce overconfidence, despite employing the same training strategy as the random baseline. The key aspect to highlight is that both methods facilitate efficient data selection with minimal computational overhead, i.e., no additional learnable parameters. Consequently, the combination of both methods shows their synergy with further improvements, resulting in the reduction of overconfidence.

        \paragraph{Mixing Location of Intermediate Feature in CMaM.}
            \Cref{tab:B} presents the results of ablation studies on CMaM, with a focus on the mixing position of intermediate features for a pair of mixed inputs. The experiments are conducted in units of residual blocks within ResNet-18 structure used as the experimental model. In this context, $I$~$=$~$1$ indicates that the mixing process occurs after passing the first residual block. In conclusion, it is observed that when the mixing occurs in a shallower layer rather than a deeper layer within the model, there is a tendency for a higher improvement in accuracy due to a further reduction in overconfidence. This is because, when mixed in a deeper layer, the parameters of the lower layers remain unaffected by the expanded training distribution resulting from the mixing process of intermediate features. In other words, this demonstrates that the extended training distribution by mixing inputs and intermediate features of a lower layer assists the model in mitigating bias. For this reason, we chose $I$~$=$~$1$ as the default option for CMaM, as it exhibits the lowest overconfidence error. All subsequent experiments will be described using this option.

        \begin{table}[!t]
        \begingroup
        \small
        \centering
        \setlength{\tabcolsep}{5pt}
        \begin{tabular}{ccccccc}
            \cmidrule[\heavyrulewidth]{1-7}
            \morecmidrules
            \cmidrule[\heavyrulewidth]{1-7}
            \multirow{2}{*}{} & \multirow{2}{*}[-.3em]{Metric} & \multicolumn{5}{c}{Cycle} \\ \cmidrule[\heavyrulewidth]{3-7} & & 2 & 4 & 6 & 8 & 10 \\ \cmidrule[\heavyrulewidth]{1-7}
            \multirow{2}{*}[-.2em]{$I$=1} 
                & Acc~{\small(\%)} \hfill $\uparrow$ & 70.0 & 82.5 & 86.4 & 88.6 & 89.6 \\ \cdotline{2-7}
                & OE~{\small (1e-3)} \hfill $\downarrow$ & \;\:1.5  & \;\:1.3  & \;\:1.3  & \;\:0.9  & \;\:0.6 \\ \cmidrule[\heavyrulewidth]{1-7}
            \multirow{2}{*}[-.2em]{$I$=2} 
                & Acc~{\small(\%)} \hfill $\uparrow$ & 71.0 & 83.4 & 86.8 & 88.7 & 89.7 \\ \cdotline{2-7}
                & OE~{\small (1e-3)} \hfill $\downarrow$ & \;\:2.9  & \;\:1.9  & \;\:1.4  & \;\:1.2  & \;\:0.7 \\ \cmidrule[\heavyrulewidth]{1-7}
            \multirow{2}{*}[-.2em]{$I$=3} 
                & Acc~{\small(\%)} \hfill $\uparrow$ & 70.5 & 81.5 & 85.3 & 87.3 & 88.4 \\ \cdotline{2-7}
                & OE~{\small (1e-3)} \hfill $\downarrow$ & 24.9 & \;\:8.8  & \;\:3.4  & \;\:2.8  & \;\:1.9 \\ \cmidrule[\heavyrulewidth]{1-7}
            \multirow{2}{*}[-.2em]{$I$=4} 
                & Acc~{\small(\%)} \hfill $\uparrow$ & 67.9 & 81.3 & 85.7 & 87.7 & 88.7 \\ \cdotline{2-7}
                & OE~{\small (1e-3)} \hfill $\downarrow$ & 20.0 & \;\:6.4  & \;\:4.8  & \;\:4.2  & \;\:1.9 \\
            \cmidrule[\heavyrulewidth]{1-7}
            \morecmidrules
            \cmidrule[\heavyrulewidth]{1-7}
        \end{tabular}
        \vspace{-2mm}
        \caption{\textbf{Ablation Study of Mixing Location in CMaM.} The number of samples budgeted per cycle is 1K. A higher accuracy and a lower OE indicate better results.\vspace{-2mm}}
        \label{tab:B}
        \endgroup
        \end{table}
            
        %%%%%%%
        % Exp C & D
        %%%%%%%   
    
        \begin{table}[!t]
            \begingroup
            \setlength{\tabcolsep}{1pt}
    	\begin{subtable}[h]{0.222\textwidth}
                \small
                \begin{tabular}{cccc}
                    \cmidrule[\heavyrulewidth]{1-4}
                    \morecmidrules
                    \cmidrule[\heavyrulewidth]{1-4}
                    \multirow{2}{*}{} & \multirow{2}{*}[-.3em]{Metric} & \multicolumn{2}{c}{Cycle} \\ \cmidrule[\heavyrulewidth]{3-4} & & 1 & 10 \\ \cmidrule[\heavyrulewidth]{1-4}
                    \multirow{2}{*}[-.2em]{OO4AL} 
                        & Acc~{\small(\%)} \hfill $\uparrow$ & 26.7 & 66.8 \\ \cdotline{2-4}
                        & OE~{\small (1e-2)} \hfill $\downarrow$ & 16.3 & \;\:0.2 \\ \cmidrule[\heavyrulewidth]{1-4}
                    \multirow{2}{*}[-.2em]{\begin{tabular}[c]{@{}c@{}}CMaM\\ only\end{tabular}}  
                        & Acc~{\small(\%)} \hfill $\uparrow$  & 26.7 & 65.9 \\ \cdotline{2-4}
                        & OE~{\small (1e-2)} \hfill $\downarrow$ & 16.3 & \;\:0.5 \\ \cmidrule[\heavyrulewidth]{1-4}
                    \multirow{2}{*}[-.2em]{\begin{tabular}[c]{@{}c@{}}Mixup\\ only\end{tabular}} 
                        & Acc~{\small(\%)} \hfill $\uparrow$ & 26.0 & 63.7 \\ \cdotline{2-4}
                        & OE~{\small (1e-2)} \hfill $\downarrow$ & 21.7 & \;\:1.8 \\
                    \cmidrule[\heavyrulewidth]{1-4}
                    \morecmidrules
                    \cmidrule[\heavyrulewidth]{1-4}
                \end{tabular}
            
    		\caption{CMaM vs.}
    		\label{tab:ExpC}
    	\end{subtable}
    	% \hspace{0.2cm}
        \setlength{\tabcolsep}{1pt}
    	\begin{subtable}[h]{0.243\textwidth}
                \small
                \begin{tabular}{cccc}
                    \cmidrule[\heavyrulewidth]{1-4}
                    \morecmidrules
                    \cmidrule[\heavyrulewidth]{1-4}
                    \multirow{2}{*}{} & \multirow{2}{*}[-.3em]{Metric} & \multicolumn{2}{c}{Cycle} \\ \cmidrule[\heavyrulewidth]{3-4} & & 1 & 10 \\ \cmidrule[\heavyrulewidth]{1-4}
                    \multirow{2}{*}[-.2em]{\begin{tabular}[c]{@{}c@{}}RankedMS\\ only\end{tabular}} 
                        & Acc~{\small(\%)} \hfill $\uparrow$ & 23.8 & 49.0 \\ \cdotline{2-4}
                        & OE~{\small (1e-2)} \hfill $\downarrow$ & 56.3 & 30.6 \\ \cmidrule[\heavyrulewidth]{1-4}
                    \multirow{2}{*}[-.2em]{\begin{tabular}[c]{@{}c@{}}Entropy\\ only\end{tabular}}  
                        & Acc~{\small(\%)} \hfill $\uparrow$  & 23.8 & 47.4 \\ \cdotline{2-4}
                        & OE~{\small (1e-2)} \hfill $\downarrow$ & 56.3 & 30.8 \\ \cmidrule[\heavyrulewidth]{1-4}
                    \multirow{2}{*}[-.2em]{\begin{tabular}[c]{@{}c@{}}Margin\\ only\end{tabular}} 
                        & Acc~{\small(\%)} \hfill $\uparrow$ & 23.8 & 48.2 \\ \cdotline{2-4}
                        & OE~{\small (1e-2)} \hfill $\downarrow$ & 56.3 & 30.7 \\
                    \cmidrule[\heavyrulewidth]{1-4}
                    \morecmidrules
                    \cmidrule[\heavyrulewidth]{1-4}
                \end{tabular}
                
                \caption{RankedMS vs.}
    		\label{tab:ExpD}
    	\end{subtable}
        \vspace{-2mm}
    	\caption{\textbf{Comparison to the Most Related Methods.} The number of samples budgeted per cycle is 0.1K. A higher accuracy and a lower OE indicate better results. \vspace{-2mm}}
    	\label{tab:ExpC_D}
            \endgroup
        \end{table}

    \subsection{Detailed Analyses}
        \paragraph{Setup and Implementation Details.}
            Experiments are conducted on a smaller budget than in the ablation studies, i.e., 0.1K or 0.5K per cycle, to better explain the effectiveness of the proposed method. All other settings are the same as in the ablation studies.
            
        \paragraph{Comparison of CMaM vs. Mixup.}
            This section provides an analysis of the differences between CMaM and the mixup within the context of active learning. CMaM was designed to mitigate bias by expanding the training distribution beyond the constraints imposed by limited training samples in the active learning scenario. So, this experiment is conducted in a smaller budget setting, i.e., 0.1K every cycle, to better demonstrate the design intent of our method. As shown in~\Cref{tab:ExpC}, a comparison is presented between the two training strategies, i.e., CMaM only vs. Mixup only, with randomly selected samples in each cycle. The results indicate that CMaM effectively calibrates the model, contributing to improved accuracy and reduced overconfident predictions as compared to the mixup. Furthermore, this suggests that CMaM has the benefit of generalizing the model by further widening the training distribution than the mixup. In conclusion, it is proven that our method is appropriately designed for its intended purpose in the active learning scenario.

            \begin{figure}[t!]
                \centering
                \includegraphics[width=0.95\linewidth]{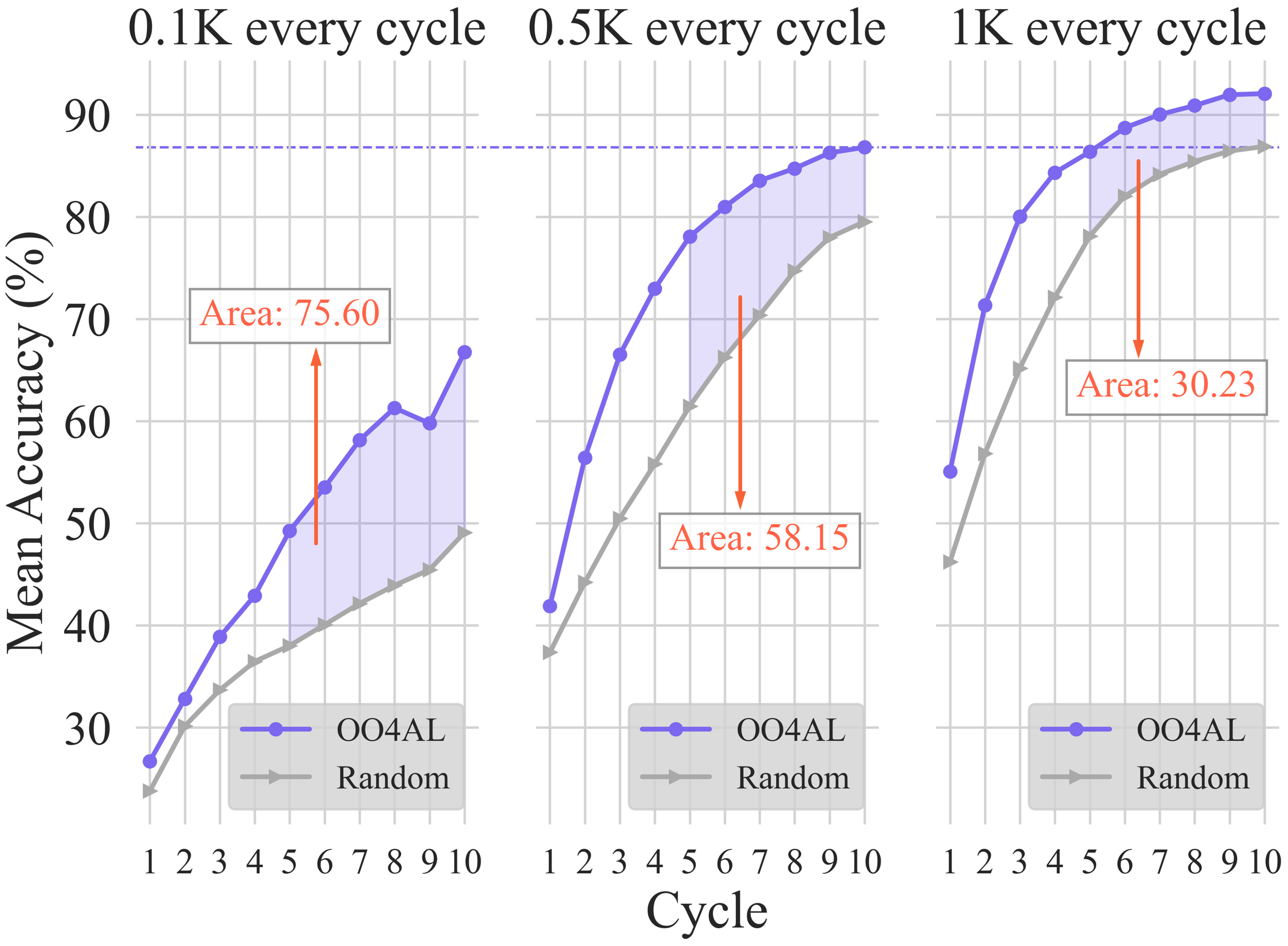} 
                \vspace{-0mm}
                \caption{\textbf{Analysis Based on Budget Size.} The number of samples budgeted per cycle is 0.1K, 0.5K, and 1K.\vspace{-4mm}} \label{fig:experimentE}
            \end{figure}

        \paragraph{Comparison of RankedMS vs. Entropy and Margin.}
            To explicitly quantify uncertainty, RankedMS leverages model predictions similar to usual approaches such as Entropy sampling and Margin sampling, but there is a difference between them in how sensitively they are designed to consider overconfidence. \Cref{tab:ExpD} offers a comparison of them in terms of alleviating overconfidence. Similar to the preceding section, the experiment is executed within a smaller budgetary framework to allow for better explanation. The three approaches exhibit relatively lower overall performance compared to~\Cref{tab:ExpC}, which directly addresses training strategies. This is because the three approaches are optimized using a conventional strategy without any extra steps to decrease overconfidence in the model; thus, the model is more vulnerable to bias in small samples. However, when compared to the other two approaches in~\Cref{tab:ExpD}, the proposed RankedMS demonstrates the best performance, suggesting that it is most sensitive to overconfidence. In addition, when it is combined with CMaM, as shown in OO4AL of~\Cref{tab:ExpC}, overconfidence can be reduced more, even in a scenario with a significantly insufficient budget. This implies that consideration of overconfidence in the quantification of uncertainty during the sampling process is one of the effective solutions.

        \paragraph{Calibration Capability Depending on Budget Size.}
            In this section, we analyze the calibration capacity of our approach across budgets of different sizes. According to the results offered in~\Cref{fig:experimentE}, our method exhibits significant enhancements across budgets when compared to the random baseline. Moreover, it is worth noting that the enhancements become more apparent in the latter half of the cycle as the budget shrinks. This indicates that the random baseline, which lacks any strategy for overconfidence, becomes increasingly biased as the number of samples decreases, whereas the proposed method successfully alleviates overconfidence in the model. Additionally, it deserves to be emphasized that the 10-cycle outcome under a 0.5K budget of the proposed strategy stands in line with the 10-cycle outcome under a 1K budget of the random baseline. This means that our strategy can result in a 50\% reduction in budget expenditures compared to a random selection of data.

            \begin{figure}[t!]
                \centering
                \includegraphics[width=0.95\linewidth]{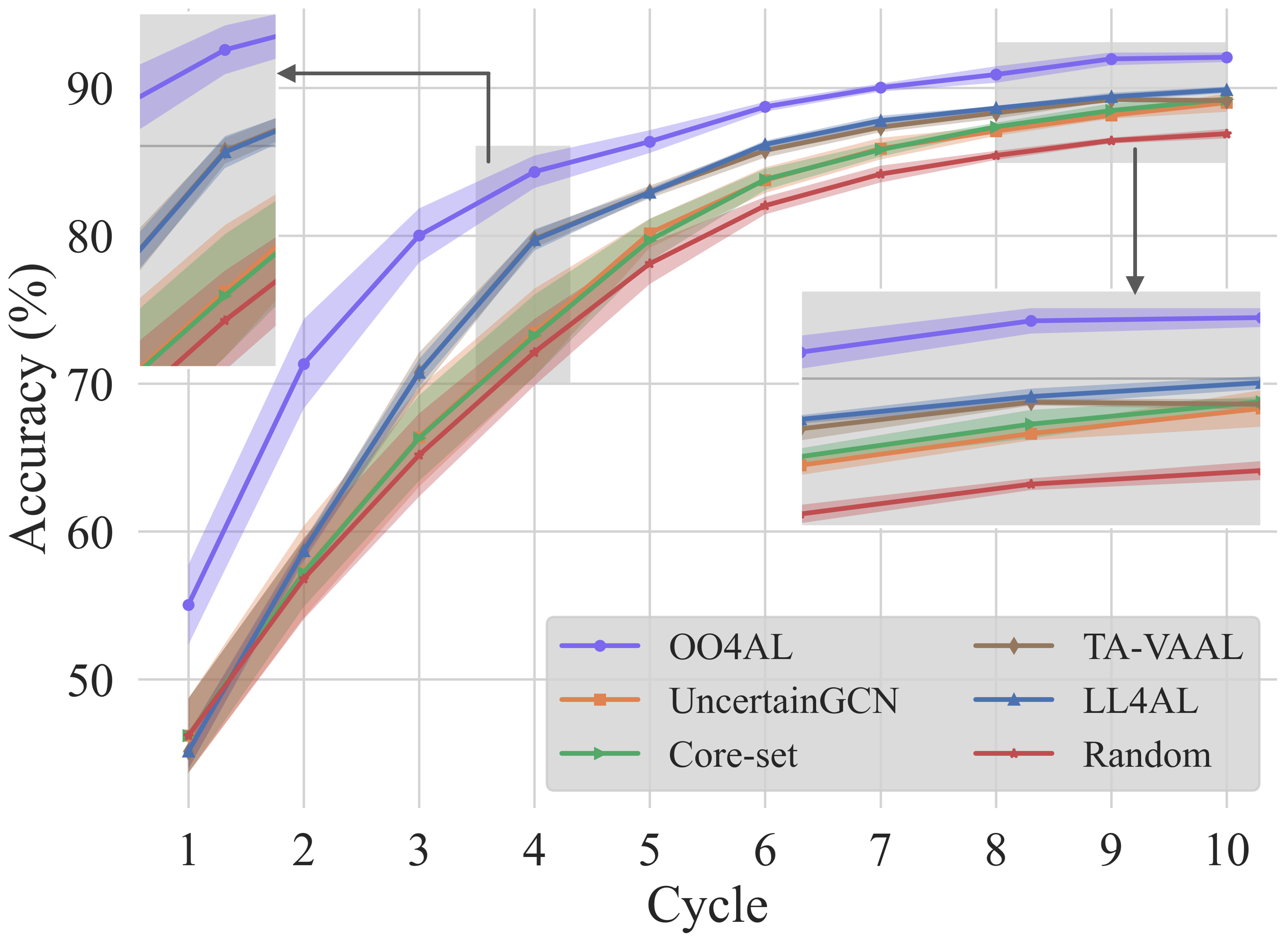} 
                \vspace{-0mm}
                \caption{\textbf{Benchmark on CIFAR10.} The number of samples budgeted per cycle is 1K. The accuracy is expressed as the mean and standard deviation of five trials. \vspace{-0mm}} \label{fig:fig_cifar10}
            \end{figure}
            \begin{figure}[t!]
                \centering
                \includegraphics[width=0.95\linewidth]{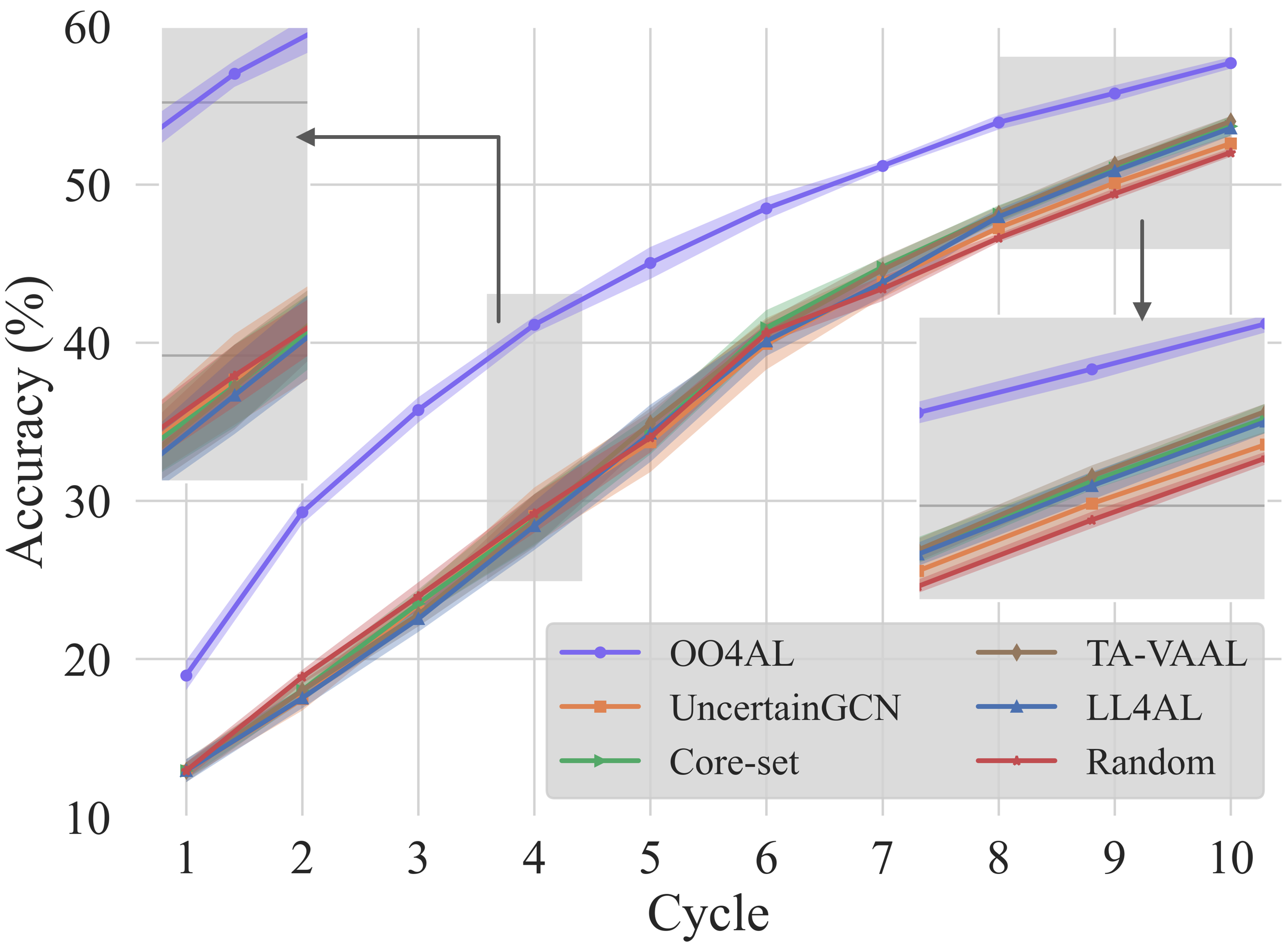} 
                \vspace{-0mm}
                \caption{\textbf{Benchmark on CIFAR100.} The number of samples budgeted per cycle is 1K. The accuracy is expressed as the mean and standard deviation of five trials. \vspace{-4mm}} \label{fig:fig_cifar100}
            \end{figure}
    \subsection{Benchmark on Standard Datasets}

        \paragraph{Setup and Implementation Details.}
            Experiments were conducted on CIFAR10 and CIFAR100~\cite{cifar}. CIFAR100 has the same size and number of sets as CIFAR10, but for 100 classes. For performance readability, we benchmarked with some of the most well-known methods in active learning: a random baseline, Core-set~\cite{coresets}, LL4AL~\cite{lloss}, UncertainGCN~\cite{gcnal}, and TA-VAAL~\cite{tavaal}. They were reimplemented with their own selection and training strategy for 1K samples per cycle; all other settings are the same for a fair comparison. These details can be found in the supplementary material.

        \paragraph{Results.}

            \Cref{fig:fig_cifar10} and~\ref{fig:fig_cifar100} show comparison results with other methods. Our method demonstrates state-of-the-art performance, which is consistent across both datasets. In particular, it is notable that although the samples randomly selected in the first cycle are equal to all methods, the proposed method exhibits significant improvements compared to other methods. This implies that the model can acquire knowledge from a broader range of the training distribution via CMaM, thereby enhancing its generalization capability. It is also consistently observed that our method surpasses others across all subsequent cycles in selecting and training strategies in its own way. These results suggest that it is crucial to address overconfidence in the active learning scenario. The supplementary material contains detailed performance numbers.

            \begin{figure}[t!]
                \centering
                \includegraphics[width=0.95\linewidth]{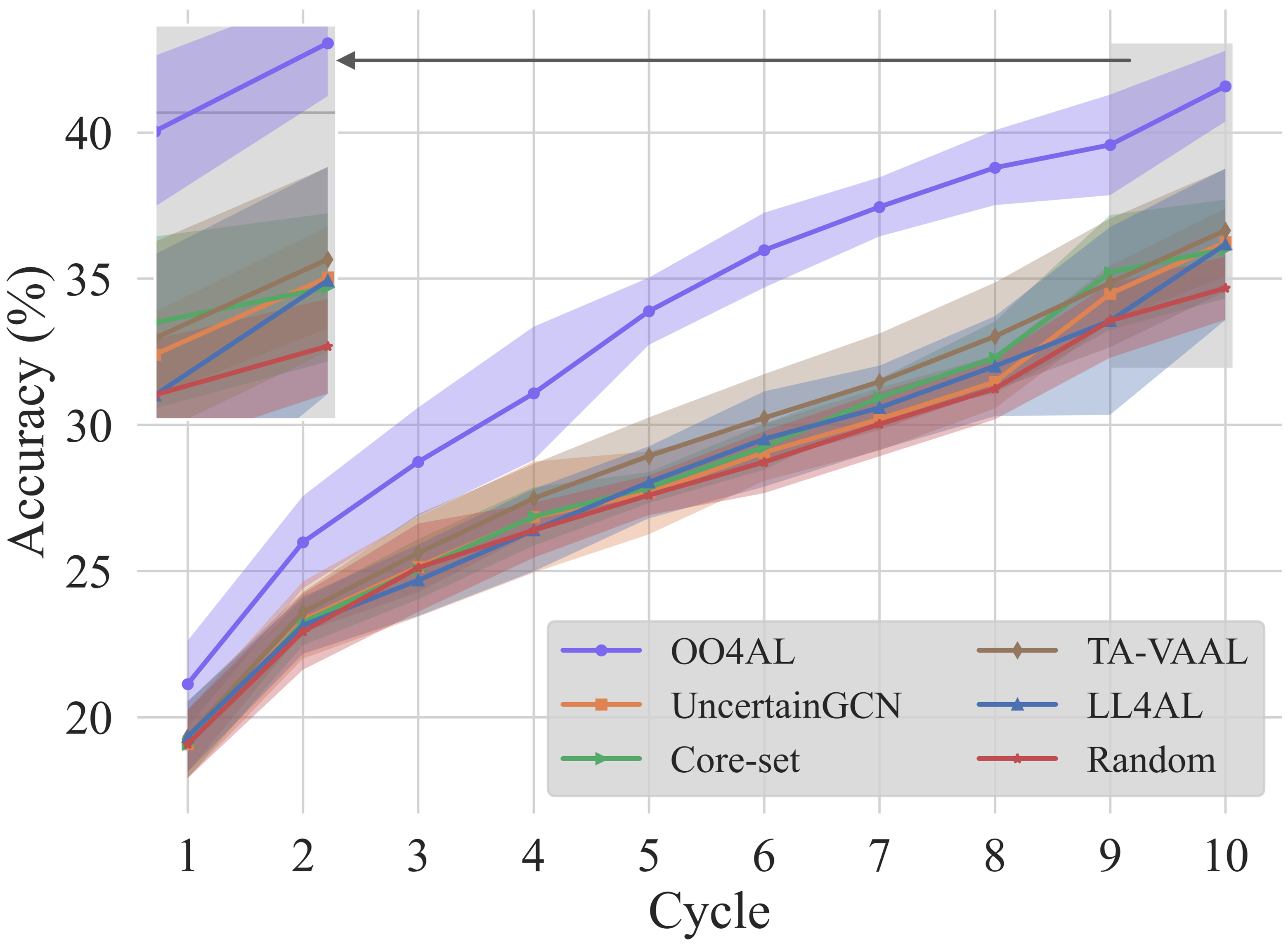} 
                \caption{\textbf{Benchmark on the imbalanced CIFAR10.} The number of samples budgeted per cycle is 0.1K. The accuracy is expressed as the mean and standard deviation of five trials. \vspace{-4mm}} \label{fig:exp_g}
            \end{figure}
            
    \subsection{Benchmark on an Imbalanced Dataset}
        \paragraph{Setup and Implementation Details.}
            Experiments were conducted on the imbalanced CIFAR10, which is the same as the form processed in the previous study~\cite{tavaal}, with five classes containing 100 times fewer samples than the other five classes. All other settings are the same as in the previous section.
        \paragraph{Results.}
            Due to the disproportionate number of samples in the dataset covered in this section, there is a high potential for the model to be biased when a limited number of samples are added in each cycle. However, upon executing the benchmark on this dataset, as depicted in~\Cref{fig:exp_g}, our method shows superior results compared to other methods. This implies that the consideration of overconfidence remains effective even within the context of an imbalance.

    \subsection{Benchmark on a Large-Scale Dataset}
        \paragraph{Setup and Implementation Details.}
            Experiments were conducted on ImageNet~\cite{imagenet}. The dataset comprises 1.2 million images for 1,000 classes. For budgets of selection, 10\% of samples are added to the labeled set in the first cycle and 5\% in the subsequent cycles. The hyperparameters for this dataset are covered in the supplemental material, and all other settings are the same as in the previous section. Due to the size of the dataset, each experiment took considerable time, so we only compared our method to the random baseline and LL4AL.

\begin{figure}[t!]
                \centering
                \includegraphics[width=0.95\linewidth]{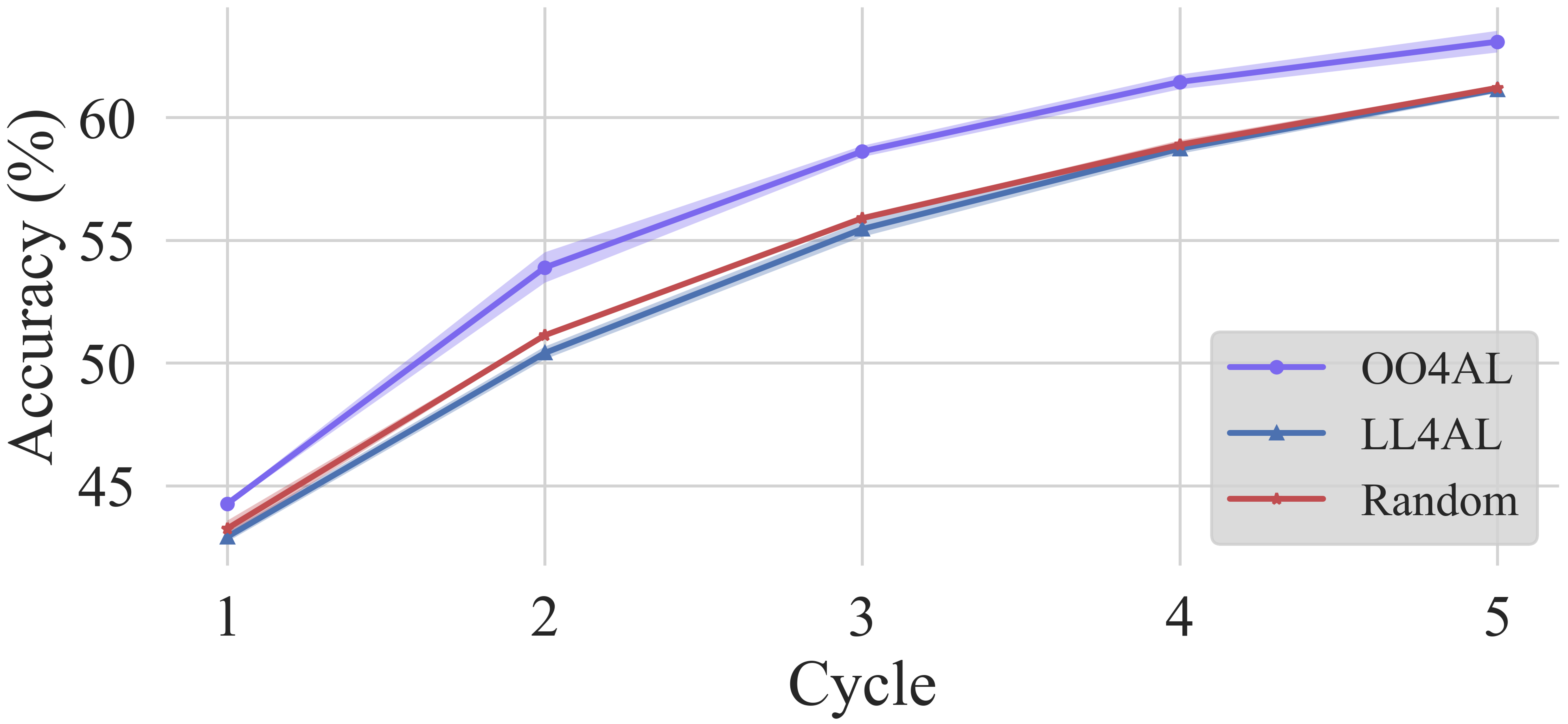} 
                \vspace{-0mm}
                \caption{\textbf{Benchmark on ImageNet.} The number of samples budgeted per cycle is 10\% for the first cycle and 5\% for the remaining cycles. The accuracy is expressed as the mean and standard deviation of three trials. \vspace{-2mm}} \label{fig:exp_h}
            \end{figure}

        \paragraph{Results.}
            As seen in~\Cref{fig:exp_h}, the proposed method yields distinct results compared to the other two methods on the large-scale dataset. This indicates that our method can be employed for effective data mining even in the real world, where massive amounts of data are handled.

\section{Conclusion}
    We proposed two novel strategies, CMaM and RankedMS, to mitigate overconfidence in the active learning scenario. The former extended the training distribution within limited samples, and the latter prevented the selection of the data from causing overconfident predictions. As a result, It was observed that valuable data can be efficiently selected with minimal computational overhead in imbalanced or large-scale setups as well as in standard setups. In other words, we proved that considering overconfidence as a significant factor is crucial for enhancing an active learning scenario.

\begin{figure}[h]
        \centering
        \includegraphics[width=0.7\linewidth]{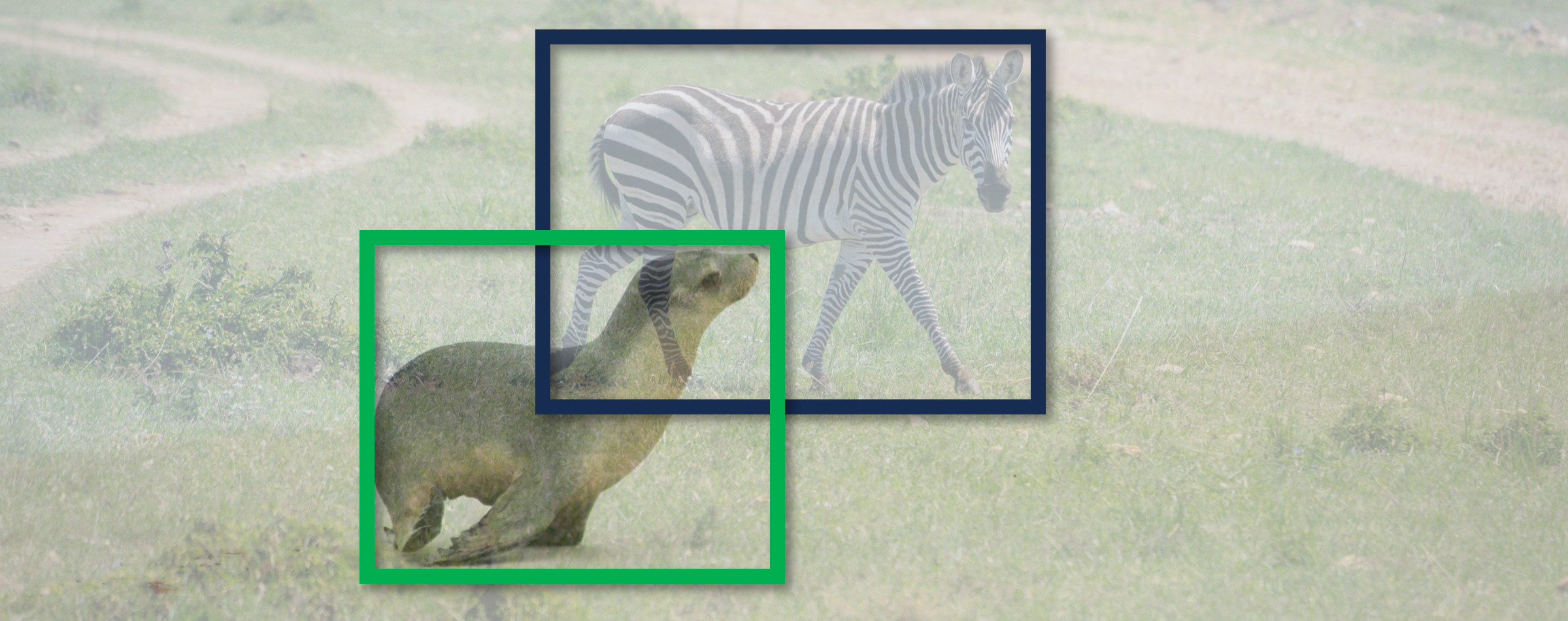} 
        \vspace{-0mm}
        \caption{\textbf{Example of Limitation.} This shows a limitation that arises when our method is applied for object detection.\vspace{-0mm}} \label{fig:conclusion}
    \end{figure}

    Here, we reveal that there is a limitation to employing our method for other tasks. In an object detection task, for instance, it is necessary to assign a single label to each bounding box. However, when our method is used, different labels should be assigned to overlapping and non-overlapping areas within a box, as shown in~\Cref{fig:conclusion}. As demonstrated in this paper, if this obstacle can be resolved in future research, it is evident that effective selection using our method will be achieved in a variety of tasks.

\bibliography{aaai24.bib}

\clearpage
\appendix

\setcounter{figure}{0}
\setcounter{equation}{0}
\setcounter{table}{0}

\renewcommand\thefigure{\Alph{figure}}
\renewcommand\theequation{\roman{equation}}
\renewcommand\thetable{\Alph{table}}

\section{Supplementary Material}

In this document, we provide supplementary information that was excluded from the main script due to space constraints. Appendix~\ref{imple} shows the implementation details of our proposed strategy. Appendix~\ref{detail_result} provides detailed numerical results for the benchmark. 

\subsection{Further Implementation Details}\label{imple}

For all experiments conducted on CIFAR10, CIFAR100, and imbalanced CIFAR10, the model was trained for 200~epochs with a batch size of 128 per cycle using the labeled dataset. It was optimized through Stochastic Gradient Descent (SGD) with a learning rate of 0.1, a momentum of 0.9, and a weight decay of 5e-4; after 160 epochs, the learning rate was decreased by one-tenth. We used the random horizontal flip and the random crop with padding as default data augmentations. For all experiments conducted on ImageNet, the model was trained for 100 epochs with a batch size of 256 per cycle using the labeled dataset. It was optimized through SGD with a learning rate of 0.1, a momentum of 0.9, and a weight decay of~1e-4; after 30, 60, and 90 epochs, the learning rate was decreased by one-tenth, respectively. We used the random horizontal flip and the random crop with padding as default data augmentations. For all datasets, hyperparameters $\lambda_1$ and $\lambda_2$ for interpolation in CMaM are drawn from the beta distribution, i.e., $\lambda_1, \lambda_2 \sim Beta(0.4, 0.4)$, and for fair comparison, this is the same in experiments with the mixup. Furthermore, all experiments were conducted on consistent computing infrastructure: a single GPU on the A6000, Ubuntu 18.04, Cuda 11.7.0 with CUDNN 8, and Pytorch 1.13.1.

\subsection{Detailed Numerical Results}\label{detail_result}
This section offers the detailed numerical results that we covered in the benchmark section of the main script. The results are presented in Tables~\ref{tab:F_cifar10},~\ref{tab:F_cifar100},~\ref{tab:G}, and \ref{tab:H_ImageNet}, which represent benchmarks on CIFAR10, CIFAR100, imbalanced CIFAR10, and ImageNet, respectively.

% %%%%%% CIFAR-10

\begin{table*}[!t]
\begingroup
    \small
    \centering
    \setlength{\tabcolsep}{1pt}
    \begin{tabular}{clllllllllll}
        \cmidrule[\heavyrulewidth]{1-12}
        \morecmidrules
        \cmidrule[\heavyrulewidth]{1-12}
        \multirow{2}{*}{} & \multicolumn{1}{c}{\multirow{2}{*}[-.3em]{Metric}} & \multicolumn{10}{c}{Cycle} \\ \cmidrule[\heavyrulewidth]{3-12} 
        & & \multicolumn{1}{c}{1} & \multicolumn{1}{c}{2} & \multicolumn{1}{c}{3} & \multicolumn{1}{c}{4} & \multicolumn{1}{c}{5} & \multicolumn{1}{c}{6} & \multicolumn{1}{c}{7} & \multicolumn{1}{c}{8} & \multicolumn{1}{c}{9} & \multicolumn{1}{c}{10} \\ \cmidrule[\heavyrulewidth]{1-12}
        \multirow{2}{*}[-.2em]{Random} 
            & Acc~{\small(\%)} \:\: $\uparrow$ & \;\:46.2{\scriptsize $\pm$\;\:2.5}& \;\:56.8{\scriptsize $\pm$\;\:2.7}& \;\:65.2{\scriptsize $\pm$\;\:2.8}& \;\:72.1{\scriptsize $\pm$\;\:2.2}& \;\:78.1{\scriptsize $\pm$\;\:1.4}& \;\:82.0{\scriptsize $\pm$\;\:0.6}& \;\:84.2{\scriptsize $\pm$\;\:0.6}& \;\:85.4{\scriptsize $\pm$\;\:0.3}& \;\:86.4{\scriptsize $\pm$\;\:0.2}& \;\:86.9{\scriptsize $\pm$\;\:0.3} \\ \cdotline{2-12}
            & OE~{\small (1e-3)}  $\downarrow$ & 348.3{\scriptsize $\pm$\;\:8.3}& 243.6{\scriptsize $\pm$15.7}& 187.8{\scriptsize $\pm$15.0}& 142.9{\scriptsize $\pm$11.6}& 111.4{\scriptsize $\pm$\;\:9.1}& \;\:85.5{\scriptsize $\pm$\;\:3.2}& \;\:73.4{\scriptsize $\pm$\;\:2.8}& \;\:66.9{\scriptsize $\pm$\;\:2.0}& \;\:63.2{\scriptsize $\pm$\;\:2.3}& \;\:64.4{\scriptsize $\pm$\;\:2.4} \\ \cmidrule[\heavyrulewidth]{1-12}
        \multirow{2}{*}[-.2em]{Core-set} 
            & Acc~{\small(\%)} \:\: $\uparrow$ & \;\:46.2{\scriptsize $\pm$\;\:2.5}& \;\:57.2{\scriptsize $\pm$\;\:2.3}& \;\:66.3{\scriptsize $\pm$\;\:2.9}& \;\:73.2{\scriptsize $\pm$\;\:2.8}& \;\:79.7{\scriptsize $\pm$\;\:1.4}& \;\:83.8{\scriptsize $\pm$\;\:0.7}& \;\:85.8{\scriptsize $\pm$\;\:0.5}& \;\:87.4{\scriptsize $\pm$\;\:0.3}& \;\:88.5{\scriptsize $\pm$\;\:0.5}& \;\:89.2{\scriptsize $\pm$\;\:0.2} \\ \cdotline{2-12}
            & OE~{\small (1e-3)}  $\downarrow$ & 348.3{\scriptsize $\pm$\;\:8.3}& 241.5{\scriptsize $\pm$11.0}& 176.4{\scriptsize $\pm$16.1}& 135.2{\scriptsize $\pm$15.2}& 104.8{\scriptsize $\pm$\;\:8.5}& \;\:76.2{\scriptsize $\pm$\;\:4.4}& \;\:67.1{\scriptsize $\pm$\;\:3.1}& \;\:59.3{\scriptsize $\pm$\;\:1.7}& \;\:53.7{\scriptsize $\pm$\;\:2.1}& \;\:52.2{\scriptsize $\pm$\;\:0.9} \\ \cmidrule[\heavyrulewidth]{1-12}
        \multirow{2}{*}[-.2em]{LL4AL} 
            & Acc~{\small(\%)} \:\: $\uparrow$ & \;\:45.1{\scriptsize $\pm$\;\:1.2}& \;\:58.7{\scriptsize $\pm$\;\:0.7}& \;\:70.7{\scriptsize $\pm$\;\:0.9}& \;\:79.7{\scriptsize $\pm$\;\:0.7}& \;\:82.9{\scriptsize $\pm$\;\:0.3}& \;\:86.2{\scriptsize $\pm$\;\:0.2}& \;\:87.8{\scriptsize $\pm$\;\:0.3}& \;\:88.6{\scriptsize $\pm$\;\:0.1}& \;\:89.4{\scriptsize $\pm$\;\:0.3}& \;\:89.9{\scriptsize $\pm$\;\:0.2} \\ \cdotline{2-12}
            & OE~{\small (1e-3)}  $\downarrow$ & 349.9{\scriptsize $\pm$\;\:5.1}& 240.6{\scriptsize $\pm$\;\:9.7}& 163.2{\scriptsize $\pm$\;\:9.0}& 111.1{\scriptsize $\pm$\;\:5.7}& \;\:92.6{\scriptsize $\pm$\;\:2.6}& \;\:71.2{\scriptsize $\pm$\;\:2.0}& \;\:64.3{\scriptsize $\pm$\;\:2.9}& \;\:61.2{\scriptsize $\pm$\;\:1.3}& \;\:58.8{\scriptsize $\pm$\;\:3.4}& \;\:59.1{\scriptsize $\pm$\;\:2.5} \\ \cmidrule[\heavyrulewidth]{1-12}
        \multirow{2}{*}[-.2em]{\begin{tabular}[c]{@{}c@{}}Uncertain\\ GCN\end{tabular}}
            & Acc~{\small(\%)} \:\: $\uparrow$ & \;\:46.2{\scriptsize $\pm$\;\:2.5}& \;\:57.3{\scriptsize $\pm$\;\:2.3}& \;\:66.4{\scriptsize $\pm$\;\:2.3}& \;\:73.4{\scriptsize $\pm$\;\:2.9}& \;\:80.2{\scriptsize $\pm$\;\:1.0}& \;\:83.8{\scriptsize $\pm$\;\:0.6}& \;\:85.9{\scriptsize $\pm$\;\:0.6}& \;\:87.1{\scriptsize $\pm$\;\:0.6}& \;\:88.1{\scriptsize $\pm$\;\:0.5}& \;\:89.0{\scriptsize $\pm$\;\:0.7} \\ \cdotline{2-12}
            & OE~{\small (1e-3)}  $\downarrow$ & 348.3{\scriptsize $\pm$\;\:8.3}& 243.7{\scriptsize $\pm$11.7}& 183.0{\scriptsize $\pm$10.1}& 139.4{\scriptsize $\pm$16.0}& 107.8{\scriptsize $\pm$\;\:5.1}& \;\:80.2{\scriptsize $\pm$\;\:2.8}& \;\:69.7{\scriptsize $\pm$\;\:5.0}& \;\:62.9{\scriptsize $\pm$\;\:3.0}& \;\:57.5{\scriptsize $\pm$\;\:2.6}& \;\:55.4{\scriptsize $\pm$\;\:4.4} \\ \cmidrule[\heavyrulewidth]{1-12}
        \multirow{2}{*}[-.2em]{TA-VAAL} 
            & Acc~{\small(\%)} \:\: $\uparrow$ & \;\:45.1{\scriptsize $\pm$\;\:1.2}& \;\:58.7{\scriptsize $\pm$\;\:0.5}& \;\:70.7{\scriptsize $\pm$\;\:1.4}& \;\:79.8{\scriptsize $\pm$\;\:0.6}& \;\:83.0{\scriptsize $\pm$\;\:0.4}& \;\:85.8{\scriptsize $\pm$\;\:0.5}& \;\:87.3{\scriptsize $\pm$\;\:0.3}& \;\:88.3{\scriptsize $\pm$\;\:0.4}& \;\:89.2{\scriptsize $\pm$\;\:0.1}& \;\:89.1{\scriptsize $\pm$\;\:0.2} \\ \cdotline{2-12}
            & OE~{\small (1e-3)}  $\downarrow$ & 349.9{\scriptsize $\pm$\;\:5.1}& 240.5{\scriptsize $\pm$\;\:8.0}& 163.9{\scriptsize $\pm$12.3}& 109.2{\scriptsize $\pm$\;\:4.5}& \;\:91.2{\scriptsize $\pm$\;\:3.4}& \;\:74.0{\scriptsize $\pm$\;\:1.9}& \;\:65.5{\scriptsize $\pm$\;\:0.3}& \;\:63.8{\scriptsize $\pm$\;\:2.2}& \;\:59.4{\scriptsize $\pm$\;\:0.4}& \;\:60.7{\scriptsize $\pm$\;\:0.6} \\ \cmidrule[\heavyrulewidth]{1-12}
        \multirow{2}{*}[-.2em]{OO4AL} 
            & Acc~{\small(\%)} \:\: $\uparrow$ & \;\:55.0{\scriptsize $\pm$\;\:2.7}& \;\:71.3{\scriptsize $\pm$\;\:3.0}& \;\:80.0{\scriptsize $\pm$\;\:1.8}& \;\:84.3{\scriptsize $\pm$\;\:1.1}& \;\:86.4{\scriptsize $\pm$\;\:0.8}& \;\:88.7{\scriptsize $\pm$\;\:0.3}& \;\:90.0{\scriptsize $\pm$\;\:0.3}& \;\:90.9{\scriptsize $\pm$\;\:0.6}& \;\:92.0{\scriptsize $\pm$\;\:0.4}& \;\:92.1{\scriptsize $\pm$\;\:0.3} \\ \cdotline{2-12}
            & OE~{\small (1e-3)}  $\downarrow$ & \;\:25.9{\scriptsize $\pm$12.1}& \;\:\;\:0.6{\scriptsize $\pm$\;\:0.2}& \;\:\;\:0.3{\scriptsize $\pm$\;\:0.2}& \;\:\;\:0.3{\scriptsize $\pm$\;\:0.1}& \;\:\;\:0.2{\scriptsize $\pm$\;\:0.1}& \;\:\;\:0.3{\scriptsize $\pm$\;\:0.2}& \;\:\;\:0.2{\scriptsize $\pm$\;\:0.2}& \;\:\;\:0.2{\scriptsize $\pm$\;\:0.1}& \;\:\;\:0.2{\scriptsize $\pm$\;\:0.1}& \;\:\;\:0.2{\scriptsize $\pm$\;\:0.1} \\ 
        \cmidrule[\heavyrulewidth]{1-12}
        \morecmidrules
        \cmidrule[\heavyrulewidth]{1-12}
    \end{tabular}
    \caption{\textbf{Benchmark on CIFAR10.} The number of samples budgeted per cycle is 1K. The accuracy and OE are expressed as the mean and standard deviation of five trials, respectively. A higher accuracy and a lower OE indicate better results.}
    \label{tab:F_cifar10}
\endgroup
\end{table*}

% %%%%%% CIFAR-100

\begin{table*}[!t]
\begingroup
    \small
    \centering
    \setlength{\tabcolsep}{1pt}
    \begin{tabular}{clllllllllll}
        \cmidrule[\heavyrulewidth]{1-12}
        \morecmidrules
        \cmidrule[\heavyrulewidth]{1-12}
        \multirow{2}{*}{} & \multicolumn{1}{c}{\multirow{2}{*}[-.3em]{Metric}} & \multicolumn{10}{c}{Cycle} \\ \cmidrule[\heavyrulewidth]{3-12} 
        & & \multicolumn{1}{c}{1} & \multicolumn{1}{c}{2} & \multicolumn{1}{c}{3} & \multicolumn{1}{c}{4} & \multicolumn{1}{c}{5} & \multicolumn{1}{c}{6} & \multicolumn{1}{c}{7} & \multicolumn{1}{c}{8} & \multicolumn{1}{c}{9} & \multicolumn{1}{c}{10} \\ \cmidrule[\heavyrulewidth]{1-12}
        \multirow{2}{*}[-.2em]{Random} 
            & Acc~{\small(\%)} \:\: $\uparrow$ & \;\:12.9{\scriptsize $\pm$\;\:0.5}& \;\:18.9{\scriptsize $\pm$\;\:0.4}& \;\:23.9{\scriptsize $\pm$\;\:0.9}& \;\:29.2{\scriptsize $\pm$\;\:1.2}& \;\:34.0{\scriptsize $\pm$\;\:0.9}& \;\:40.6{\scriptsize $\pm$\;\:0.6}& \;\:43.4{\scriptsize $\pm$\;\:0.8}& \;\:46.6{\scriptsize $\pm$\;\:0.3}& \;\:49.4{\scriptsize $\pm$\;\:0.3}& \;\:52.0{\scriptsize $\pm$\;\:0.3} \\ \cdotline{2-12}
            & OE~{\small (1e-2)}  $\downarrow$ & \;\:30.4{\scriptsize $\pm$\;\:1.0}& \;\:13.9{\scriptsize $\pm$\;\:1.0}& \;\:21.0{\scriptsize $\pm$\;\:3.3}& \;\:19.7{\scriptsize $\pm$\;\:3.5}& \;\:23.4{\scriptsize $\pm$\;\:0.5}& \;\:13.2{\scriptsize $\pm$\;\:3.9}& \;\:15.8{\scriptsize $\pm$\;\:1.5}& \;\:14.6{\scriptsize $\pm$\;\:0.3}& \;\:13.5{\scriptsize $\pm$\;\:0.3}& \;\:13.6{\scriptsize $\pm$\;\:0.4} \\ \cmidrule[\heavyrulewidth]{1-12}
        \multirow{2}{*}[-.2em]{Core-set} 
            & Acc~{\small(\%)} \:\: $\uparrow$ & \;\:12.9{\scriptsize $\pm$\;\:0.5}& \;\:18.0{\scriptsize $\pm$\;\:0.7}& \;\:23.5{\scriptsize $\pm$\;\:0.9}& \;\:28.8{\scriptsize $\pm$\;\:1.6}& \;\:34.2{\scriptsize $\pm$\;\:1.2}& \;\:40.9{\scriptsize $\pm$\;\:1.1}& \;\:44.8{\scriptsize $\pm$\;\:0.6}& \;\:48.1{\scriptsize $\pm$\;\:0.6}& \;\:51.0{\scriptsize $\pm$\;\:0.4}& \;\:53.7{\scriptsize $\pm$\;\:0.6} \\ \cdotline{2-12}
            & OE~{\small (1e-2)}  $\downarrow$ & \;\:30.4{\scriptsize $\pm$\;\:1.0}& \;\:19.4{\scriptsize $\pm$\;\:5.9}& \;\:17.8{\scriptsize $\pm$\;\:4.5}& \;\:18.4{\scriptsize $\pm$\;\:4.8}& \;\:22.3{\scriptsize $\pm$\;\:0.5}& \;\:14.3{\scriptsize $\pm$\;\:5.0}& \;\:13.4{\scriptsize $\pm$\;\:2.3}& \;\:13.4{\scriptsize $\pm$\;\:0.2}& \;\:12.4{\scriptsize $\pm$\;\:0.3}& \;\:12.2{\scriptsize $\pm$\;\:0.4} \\ \cmidrule[\heavyrulewidth]{1-12}
        \multirow{2}{*}[-.2em]{LL4AL} 
            & Acc~{\small(\%)} \:\: $\uparrow$ & \;\:12.9{\scriptsize $\pm$\;\:0.7}& \;\:17.5{\scriptsize $\pm$\;\:0.6}& \;\:22.5{\scriptsize $\pm$\;\:0.8}& \;\:28.4{\scriptsize $\pm$\;\:1.5}& \;\:34.3{\scriptsize $\pm$\;\:1.8}& \;\:40.1{\scriptsize $\pm$\;\:0.9}& \;\:43.8{\scriptsize $\pm$\;\:0.9}& \;\:48.0{\scriptsize $\pm$\;\:0.5}& \;\:50.8{\scriptsize $\pm$\;\:0.6}& \;\:53.5{\scriptsize $\pm$\;\:0.5} \\ \cdotline{2-12}
            & OE~{\small (1e-2)}  $\downarrow$ & \;\:30.9{\scriptsize $\pm$\;\:0.5}& \;\:18.3{\scriptsize $\pm$\;\:0.8}& \;\:13.2{\scriptsize $\pm$\;\:0.8}& \;\:15.5{\scriptsize $\pm$\;\:4.4}& \;\:25.8{\scriptsize $\pm$\;\:0.6}& \;\:19.8{\scriptsize $\pm$\;\:0.8}& \;\:20.0{\scriptsize $\pm$\;\:0.7}& \;\:18.0{\scriptsize $\pm$\;\:0.3}& \;\:16.8{\scriptsize $\pm$\;\:0.3}& \;\:16.5{\scriptsize $\pm$\;\:0.3} \\ \cmidrule[\heavyrulewidth]{1-12}
        \multirow{2}{*}[-.2em]{\begin{tabular}[c]{@{}c@{}}Uncertain\\ GCN\end{tabular}}
            & Acc~{\small(\%)} \:\: $\uparrow$ & \;\:12.9{\scriptsize $\pm$\;\:0.5}& \;\:17.5{\scriptsize $\pm$\;\:0.7}& \;\:23.4{\scriptsize $\pm$\;\:0.8}& \;\:29.0{\scriptsize $\pm$\;\:1.8}& \;\:33.7{\scriptsize $\pm$\;\:1.9}& \;\:39.9{\scriptsize $\pm$\;\:1.6}& \;\:43.8{\scriptsize $\pm$\;\:1.0}& \;\:47.2{\scriptsize $\pm$\;\:0.8}& \;\:50.1{\scriptsize $\pm$\;\:0.6}& \;\:52.6{\scriptsize $\pm$\;\:0.6} \\ \cdotline{2-12}
            & OE~{\small (1e-2)}  $\downarrow$ & \;\:30.4{\scriptsize $\pm$\;\:1.0}& \;\:22.1{\scriptsize $\pm$\;\:4.8}& \;\:21.2{\scriptsize $\pm$\;\:4.9}& \;\:17.4{\scriptsize $\pm$\;\:3.0}& \;\:25.1{\scriptsize $\pm$\;\:1.0}& \;\:19.5{\scriptsize $\pm$\;\:1.2}& \;\:18.3{\scriptsize $\pm$\;\:0.7}& \;\:16.7{\scriptsize $\pm$\;\:0.2}& \;\:15.7{\scriptsize $\pm$\;\:0.3}& \;\:15.4{\scriptsize $\pm$\;\:0.5} \\ \cmidrule[\heavyrulewidth]{1-12}
        \multirow{2}{*}[-.2em]{TA-VAAL} 
            & Acc~{\small(\%)} \:\: $\uparrow$ & \;\:12.9{\scriptsize $\pm$\;\:0.7}& \;\:18.0{\scriptsize $\pm$\;\:0.6}& \;\:22.8{\scriptsize $\pm$\;\:0.8}& \;\:28.7{\scriptsize $\pm$\;\:1.7}& \;\:34.9{\scriptsize $\pm$\;\:1.0}& \;\:40.6{\scriptsize $\pm$\;\:0.9}& \;\:44.6{\scriptsize $\pm$\;\:0.8}& \;\:48.1{\scriptsize $\pm$\;\:0.5}& \;\:51.2{\scriptsize $\pm$\;\:0.5}& \;\:54.0{\scriptsize $\pm$\;\:0.3} \\ \cdotline{2-12}
            & OE~{\small (1e-2)}  $\downarrow$ & \;\:30.9{\scriptsize $\pm$\;\:0.5}& \;\:18.5{\scriptsize $\pm$\;\:0.6}& \;\:13.1{\scriptsize $\pm$\;\:0.8}& \;\:12.5{\scriptsize $\pm$\;\:1.9}& \;\:25.4{\scriptsize $\pm$\;\:0.5}& \;\:18.5{\scriptsize $\pm$\;\:1.7}& \;\:18.6{\scriptsize $\pm$\;\:0.9}& \;\:17.4{\scriptsize $\pm$\;\:0.6}& \;\:16.3{\scriptsize $\pm$\;\:0.5}& \;\:16.1{\scriptsize $\pm$\;\:0.5} \\ \cmidrule[\heavyrulewidth]{1-12}
        \multirow{2}{*}[-.2em]{OO4AL} 
            & Acc~{\small(\%)} \:\: $\uparrow$ & \;\:18.3{\scriptsize $\pm$\;\:0.7}& \;\:28.2{\scriptsize $\pm$\;\:0.9}& \;\:36.1{\scriptsize $\pm$\;\:0.9}& \;\:41.1{\scriptsize $\pm$\;\:0.5}& \;\:45.5{\scriptsize $\pm$\;\:0.3}& \;\:50.1{\scriptsize $\pm$\;\:0.9}& \;\:52.7{\scriptsize $\pm$\;\:0.3}& \;\:54.4{\scriptsize $\pm$\;\:0.4}& \;\:56.8{\scriptsize $\pm$\;\:0.6}& \;\:58.0{\scriptsize $\pm$\;\:0.8} \\ \cdotline{2-12}
            & OE~{\small (1e-2)}  $\downarrow$ & \;\:\;\:4.1{\scriptsize $\pm$\;\:0.5}& \;\:\;\:0.8{\scriptsize $\pm$\;\:0.3}& \;\:\;\:0.3{\scriptsize $\pm$\;\:0.1}& \;\:\;\:0.2{\scriptsize $\pm$\;\:0.1}& \;\:\;\:0.2{\scriptsize $\pm$\;\:0.1}& \;\:\;\:0.1{\scriptsize $\pm$\;\:0.1}& \;\:\;\:0.1{\scriptsize $\pm$\;\:0.0}& \;\:\;\:0.1{\scriptsize $\pm$\;\:0.1}& \;\:\;\:0.1{\scriptsize $\pm$\;\:0.0}& \;\:\;\:0.1{\scriptsize $\pm$\;\:0.1} \\
        \cmidrule[\heavyrulewidth]{1-12}
        \morecmidrules
        \cmidrule[\heavyrulewidth]{1-12}
    \end{tabular}
    \caption{\textbf{Benchmark on CIFAR100. } The number of samples budgeted per cycle is 1K. The accuracy and OE are expressed as the mean and standard deviation of five trials, respectively. A higher accuracy and a lower OE indicate better results.}
    \label{tab:F_cifar100}
\endgroup
\end{table*}

% %%%%%% Imbalanced CIFAR-10

\begin{table*}[!t]
\begingroup
    \small
    \centering
    \setlength{\tabcolsep}{1pt}
    \begin{tabular}{clllllllllll}
        \cmidrule[\heavyrulewidth]{1-12}
        \morecmidrules
        \cmidrule[\heavyrulewidth]{1-12}
        \multirow{2}{*}{} & \multicolumn{1}{c}{\multirow{2}{*}[-.3em]{Metric}} & \multicolumn{10}{c}{Cycle} \\ \cmidrule[\heavyrulewidth]{3-12} 
        & & \multicolumn{1}{c}{1} & \multicolumn{1}{c}{2} & \multicolumn{1}{c}{3} & \multicolumn{1}{c}{4} & \multicolumn{1}{c}{5} & \multicolumn{1}{c}{6} & \multicolumn{1}{c}{7} & \multicolumn{1}{c}{8} & \multicolumn{1}{c}{9} & \multicolumn{1}{c}{10} \\ \cmidrule[\heavyrulewidth]{1-12}
        \multirow{2}{*}[-.2em]{Random} 
            & Acc~{\small(\%)} \:\: $\uparrow$ & \;\:19.1{\scriptsize $\pm$\;\:1.2}& \;\:22.9{\scriptsize $\pm$\;\:1.3}& \;\:25.1{\scriptsize $\pm$\;\:1.5}& \;\:26.4{\scriptsize $\pm$\;\:0.9}& \;\:27.6{\scriptsize $\pm$\;\:0.7}& \;\:28.7{\scriptsize $\pm$\;\:1.1}& \;\:30.0{\scriptsize $\pm$\;\:1.1}& \;\:31.2{\scriptsize $\pm$\;\:1.1}& \;\:33.6{\scriptsize $\pm$\;\:1.3}& \;\:34.7{\scriptsize $\pm$\;\:1.1} \\ \cdotline{2-12}
            & OE~{\small (1e-2)}  $\downarrow$ & \;\:61.8{\scriptsize $\pm$\;\:2.2}& \;\:59.2{\scriptsize $\pm$\;\:1.9}& \;\:57.2{\scriptsize $\pm$\;\:0.9}& \;\:52.6{\scriptsize $\pm$\;\:0.3}& \;\:55.9{\scriptsize $\pm$\;\:0.8}& \;\:54.4{\scriptsize $\pm$\;\:1.0}& \;\:51.9{\scriptsize $\pm$\;\:1.1}& \;\:50.3{\scriptsize $\pm$\;\:1.2}& \;\:41.6{\scriptsize $\pm$\;\:0.6}& \;\:48.5{\scriptsize $\pm$\;\:0.5} \\ \cmidrule[\heavyrulewidth]{1-12} 
        \multirow{2}{*}[-.2em]{Core-set} 
            & Acc~{\small(\%)} \:\: $\uparrow$ & \;\:19.1{\scriptsize $\pm$\;\:1.2}& \;\:23.2{\scriptsize $\pm$\;\:0.8}& \;\:25.1{\scriptsize $\pm$\;\:1.0}& \;\:26.9{\scriptsize $\pm$\;\:1.0}& \;\:27.9{\scriptsize $\pm$\;\:0.5}& \;\:29.2{\scriptsize $\pm$\;\:0.8}& \;\:30.9{\scriptsize $\pm$\;\:0.6}& \;\:32.3{\scriptsize $\pm$\;\:1.2}& \;\:35.2{\scriptsize $\pm$\;\:2.0}& \;\:36.0{\scriptsize $\pm$\;\:1.7} \\ \cdotline{2-12}
            & OE~{\small (1e-2)}  $\downarrow$ & \;\:61.8{\scriptsize $\pm$\;\:2.2}& \;\:60.6{\scriptsize $\pm$\;\:1.6}& \;\:57.4{\scriptsize $\pm$\;\:2.1}& \;\:50.8{\scriptsize $\pm$\;\:1.8}& \;\:55.8{\scriptsize $\pm$\;\:0.8}& \;\:53.4{\scriptsize $\pm$\;\:0.7}& \;\:51.2{\scriptsize $\pm$\;\:1.0}& \;\:48.4{\scriptsize $\pm$\;\:1.0}& \;\:39.0{\scriptsize $\pm$\;\:2.5}& \;\:47.4{\scriptsize $\pm$\;\:1.8} \\ \cmidrule[\heavyrulewidth]{1-12}
        \multirow{2}{*}[-.2em]{LL4AL} 
            & Acc~{\small(\%)} \:\: $\uparrow$ & \;\:19.3{\scriptsize $\pm$\;\:1.2}& \;\:23.1{\scriptsize $\pm$\;\:1.0}& \;\:24.7{\scriptsize $\pm$\;\:1.2}& \;\:26.4{\scriptsize $\pm$\;\:1.4}& \;\:28.0{\scriptsize $\pm$\;\:1.2}& \;\:29.5{\scriptsize $\pm$\;\:1.6}& \;\:30.6{\scriptsize $\pm$\;\:1.4}& \;\:32.0{\scriptsize $\pm$\;\:1.7}& \;\:33.5{\scriptsize $\pm$\;\:3.2}& \;\:36.2{\scriptsize $\pm$\;\:2.6} \\ \cdotline{2-12}
            & OE~{\small (1e-2)}  $\downarrow$ & \;\:59.0{\scriptsize $\pm$\;\:4.4}& \;\:57.8{\scriptsize $\pm$\;\:1.4}& \;\:57.0{\scriptsize $\pm$\;\:1.8}& \;\:52.8{\scriptsize $\pm$\;\:1.8}& \;\:54.8{\scriptsize $\pm$\;\:1.3}& \;\:53.9{\scriptsize $\pm$\;\:1.4}& \;\:53.1{\scriptsize $\pm$\;\:1.3}& \;\:51.7{\scriptsize $\pm$\;\:1.6}& \;\:41.9{\scriptsize $\pm$\;\:5.2}& \;\:48.3{\scriptsize $\pm$\;\:1.7} \\ \cmidrule[\heavyrulewidth]{1-12}
        \multirow{2}{*}[-.2em]{\begin{tabular}[c]{@{}c@{}}Uncertain\\ GCN\end{tabular}}
            & Acc~{\small(\%)} \:\: $\uparrow$ & \;\:19.1{\scriptsize $\pm$\;\:1.2}& \;\:23.3{\scriptsize $\pm$\;\:1.3}& \;\:25.1{\scriptsize $\pm$\;\:1.7}& \;\:26.8{\scriptsize $\pm$\;\:1.9}& \;\:27.7{\scriptsize $\pm$\;\:1.4}& \;\:29.1{\scriptsize $\pm$\;\:1.0}& \;\:30.2{\scriptsize $\pm$\;\:1.1}& \;\:31.5{\scriptsize $\pm$\;\:0.9}& \;\:34.5{\scriptsize $\pm$\;\:1.0}& \;\:36.2{\scriptsize $\pm$\;\:1.2} \\ \cdotline{2-12}
            & OE~{\small (1e-2)}  $\downarrow$ & \;\:61.8{\scriptsize $\pm$\;\:2.2}& \;\:60.2{\scriptsize $\pm$\;\:1.4}& \;\:57.7{\scriptsize $\pm$\;\:1.7}& \;\:53.1{\scriptsize $\pm$\;\:2.0}& \;\:56.4{\scriptsize $\pm$\;\:1.7}& \;\:54.6{\scriptsize $\pm$\;\:1.6}& \;\:52.9{\scriptsize $\pm$\;\:1.5}& \;\:50.7{\scriptsize $\pm$\;\:1.7}& \;\:40.1{\scriptsize $\pm$\;\:3.5}& \;\:48.4{\scriptsize $\pm$\;\:1.6} \\ \cmidrule[\heavyrulewidth]{1-12}
        \multirow{2}{*}[-.2em]{TA-VAAL}
            & Acc~{\small(\%)} \:\: $\uparrow$ & \;\:19.3{\scriptsize $\pm$\;\:1.2}& \;\:23.6{\scriptsize $\pm$\;\:0.7}& \;\:25.6{\scriptsize $\pm$\;\:1.3}& \;\:27.5{\scriptsize $\pm$\;\:1.2}& \;\:28.9{\scriptsize $\pm$\;\:1.3}& \;\:30.2{\scriptsize $\pm$\;\:1.5}& \;\:31.5{\scriptsize $\pm$\;\:1.6}& \;\:33.0{\scriptsize $\pm$\;\:1.8}& \;\:34.9{\scriptsize $\pm$\;\:2.2}& \;\:36.6{\scriptsize $\pm$\;\:2.1} \\ \cdotline{2-12}
            & OE~{\small (1e-2)}  $\downarrow$ & \;\:59.0{\scriptsize $\pm$\;\:4.4}& \;\:57.3{\scriptsize $\pm$\;\:2.9}& \;\:55.9{\scriptsize $\pm$\;\:1.3}& \;\:50.5{\scriptsize $\pm$\;\:1.8}& \;\:53.5{\scriptsize $\pm$\;\:1.3}& \;\:53.5{\scriptsize $\pm$\;\:1.2}& \;\:52.1{\scriptsize $\pm$\;\:1.2}& \;\:50.8{\scriptsize $\pm$\;\:1.5}& \;\:40.2{\scriptsize $\pm$\;\:2.8}& \;\:47.7{\scriptsize $\pm$\;\:1.5} \\ \cmidrule[\heavyrulewidth]{1-12}
        \multirow{2}{*}[-.2em]{OO4AL} 
            & Acc~{\small(\%)} \:\: $\uparrow$ & \;\:21.1{\scriptsize $\pm$\;\:1.5}& \;\:26.0{\scriptsize $\pm$\;\:1.6}& \;\:28.7{\scriptsize $\pm$\;\:1.9}& \;\:31.1{\scriptsize $\pm$\;\:2.3}& \;\:33.9{\scriptsize $\pm$\;\:1.2}& \;\:36.0{\scriptsize $\pm$\;\:1.3}& \;\:37.5{\scriptsize $\pm$\;\:1.0}& \;\:38.8{\scriptsize $\pm$\;\:1.3}& \;\:39.6{\scriptsize $\pm$\;\:1.7}& \;\:41.6{\scriptsize $\pm$\;\:1.2} \\ \cdotline{2-12}
            & OE~{\small (1e-2)}  $\downarrow$ & \;\:25.0{\scriptsize $\pm$\;\:3.5}& \;\:28.8{\scriptsize $\pm$\;\:1.3}& \;\:26.1{\scriptsize $\pm$\;\:3.3}& \;\:23.8{\scriptsize $\pm$\;\:2.0}& \;\:20.9{\scriptsize $\pm$\;\:1.3}& \;\:19.4{\scriptsize $\pm$\;\:2.0}& \;\:18.5{\scriptsize $\pm$\;\:2.5}& \;\:16.6{\scriptsize $\pm$\;\:0.7}& \;\:17.7{\scriptsize $\pm$\;\:1.7}& \;\:15.6{\scriptsize $\pm$\;\:0.7} \\ 
        \cmidrule[\heavyrulewidth]{1-12}
        \morecmidrules
        \cmidrule[\heavyrulewidth]{1-12}
    \end{tabular}
    \caption{\textbf{Benchmark on the imbalanced CIFAR10.} The number of samples budgeted per cycle is 0.1K. The accuracy and OE are expressed as the mean and standard deviation of five trials, respectively. A higher accuracy and a lower OE indicate better results.}
    \label{tab:G}
\endgroup
\end{table*}

% %%%%%% ImageNet

\begin{table*}[!t]
\begingroup
    \small
    \centering
    \setlength{\tabcolsep}{5pt}
    \begin{tabular}{cllllll}
        \cmidrule[\heavyrulewidth]{1-7}
        \morecmidrules
        \cmidrule[\heavyrulewidth]{1-7}
        \multirow{2}{*}{} & \multicolumn{1}{c}{\multirow{2}{*}[-.3em]{Metric}} & \multicolumn{5}{c}{Cycle} \\ \cmidrule[\heavyrulewidth]{3-7} 
        & & \multicolumn{1}{c}{1} & \multicolumn{1}{c}{2} & \multicolumn{1}{c}{3} & \multicolumn{1}{c}{4} & \multicolumn{1}{c}{5} \\ \cmidrule[\heavyrulewidth]{1-7}
        \multirow{2}{*}[-.2em]{Random} 
            & Acc~{\small(\%)} \:\: $\uparrow$ & \;\:\;\:43.3{\scriptsize $\pm$\;\:0.3}& \;\:\;\:51.1{\scriptsize $\pm$\;\:0.1}& \;\:\;\:55.9{\scriptsize $\pm$\;\:0.1}& \;\:\;\:58.9{\scriptsize $\pm$\;\:0.2}& \;\:\;\:61.2{\scriptsize $\pm$\;\:0.1} \\ \cdotline{2-7}
            & OE~{\small (1e-4)}  $\downarrow$ & 1075.9{\scriptsize $\pm$16.7}& 1008.8{\scriptsize $\pm$10.5}& \;\:865.5{\scriptsize $\pm$17.9}& \;\:745.6{\scriptsize $\pm$\;\:7.7}& \;\:651.6{\scriptsize $\pm$16.9} \\ \cmidrule[\heavyrulewidth]{1-7}
        \multirow{2}{*}[-.2em]{LL4AL} 
            & Acc~{\small(\%)} \:\: $\uparrow$ & \;\:\;\:42.9{\scriptsize $\pm$\;\:0.2}& \;\:\;\:50.4{\scriptsize $\pm$\;\:0.3}& \;\:\;\:55.5{\scriptsize $\pm$\;\:0.3}& \;\:\;\:58.7{\scriptsize $\pm$\;\:0.2}& \;\:\;\:61.1{\scriptsize $\pm$\;\:0.1} \\ \cdotline{2-7}
            & OE~{\small (1e-4)}  $\downarrow$ & \;\:894.1{\scriptsize $\pm$14.4}& \;\:928.9{\scriptsize $\pm$17.9}& \;\:806.4{\scriptsize $\pm$18.1}& \;\:696.9{\scriptsize $\pm$17.9}& \;\:593.6{\scriptsize $\pm$12.8} \\ \cmidrule[\heavyrulewidth]{1-7}
        \multirow{2}{*}[-.2em]{OO4AL} 
            & Acc~{\small(\%)} \:\: $\uparrow$ & \;\:\;\:44.3{\scriptsize $\pm$\;\:0.1}& \;\:\;\:53.9{\scriptsize $\pm$\;\:0.6}& \;\:\;\:58.6{\scriptsize $\pm$\;\:0.2}& \;\:\;\:61.5{\scriptsize $\pm$\;\:0.3}& \;\:\;\:63.1{\scriptsize $\pm$\;\:0.4} \\ \cdotline{2-7}
            & OE~{\small (1e-4)}  $\downarrow$ & \;\:\;\:\;\:0.3{\scriptsize $\pm$\;\:0.4}& \;\:\;\:\;\:0.2{\scriptsize $\pm$\;\:0.2}& \;\:\;\:\;\:0.3{\scriptsize $\pm$\;\:0.2}& \;\:\;\:\;\:1.0{\scriptsize $\pm$\;\:0.5}& \;\:\;\:\;\:0.6{\scriptsize $\pm$\;\:0.8} \\
        \cmidrule[\heavyrulewidth]{1-7}
        \morecmidrules
        \cmidrule[\heavyrulewidth]{1-7}
    \end{tabular}
    \caption{\textbf{Benchmark on ImageNet} The number of samples budgeted per cycle is 10\% for the first cycle and 5\% for the remaining cycles. The accuracy and OE are expressed as the mean and standard deviation of three trials, respectively. A higher accuracy and a lower OE indicate better results.}
    \label{tab:H_ImageNet}
\endgroup
\end{table*}

\end{document}